\RequirePackage{fix-cm}

\documentclass[smallextended]{svjour3}          % onecolumn (second format)

\smartqed  % flush right qed marks, e.g. at end of proof

\usepackage{graphicx}
\usepackage[figuresright]{rotating}
\usepackage{multirow}
\usepackage{authblk,amssymb,epstopdf,lineno,natbib,subfigure,color,amsmath}
\usepackage{booktabs,marvosym}
\usepackage{lscape}
\usepackage{longtable}
\usepackage{algorithm,algorithmic}
\usepackage{soul}
\usepackage{makecell}

\newcommand{\Kbf}{\mathbf K}

\newcommand{\xbf}{\mathbf x}
\newcommand{\hbf}{\mathbf h}
\newcommand{\mbf}{\mathbf m}
\newcommand{\vbf}{\mathbf v}
\newcommand{\zbf}{\mathbf z}
\newcommand{\thetabf}{\boldsymbol{\theta}}

\journalname{ }

\begin{document}

\title{Deep learning architectures for data-driven damage detection in nonlinear dynamic systems}
\titlerunning{Deep learning architectures for data-driven damage detection}        % if too long for running head

\author{Harrish Joseph         \and
           Giuseppe Quaranta \and
           Biagio Carboni \and
           Walter Lacarbonara
}

\institute{
              H. Joseph, B. Carboni, G. Quaranta, W. Lacarbonara \at
              Department of Structural and Geotechnical Engineering, \\
		Sapienza University of Rome, Via Eudossiana 18,\\
              00184 Rome, Italy 
}

\date{Received: date / Accepted: date}

\maketitle

%%%%%%%%%%%%%%%%%%%%%%%%%%%%%%%%%%
\begin{abstract}

The primary goal of structural health monitoring is to detect damage at its onset before it reaches a critical level. The in-depth investigation in the present work addresses deep learning applied to data-driven damage detection in nonlinear dynamic systems. In particular, autoencoders (AEs) and generative adversarial networks  (GANs) are implemented leveraging on 1D convolutional neural networks. The onset of damage   is detected in the investigated  nonlinear dynamic systems by exciting random vibrations of varying intensity, without prior knowledge of the system or the excitation and in unsupervised manner. The comprehensive numerical study is conducted on dynamic systems exhibiting different types of nonlinear behavior. An experimental application related to a magneto-elastic nonlinear system is also presented to corroborate the conclusions.

\end{abstract}

%%%%%%%%%%%%%%%%%%%%%%%%%%%%%%%%%%
\setcounter{tocdepth}{3}
\setcounter{secnumdepth}{3}

%%%%%%%%%%%%%%%%%%%%%%%%%%%%%%%%%%
\section{Introduction}\label{sec:intro}
Damage is any unintended alteration in material and/or geometric properties that may impact a structural system, including its boundary and connectivity conditions, leading to detrimental effects on both current and future performance \citep[e.g., ][]{worden2007fundamental}. Early damage detection is crucial for preserving reliability and performance of structural systems throughout their lifespan by enabling timely proactive maintenance. Vibration-based methods are particularly attractive for this task because they can detect damage analyzing the system dynamic response, also when damage is not severe enough to be visually apparent. The feasibility of vibration-based damage detection methods has been extensively investigated, yet most efforts so far have focused on dynamic systems exhibiting linear behavior. Carrying out a complete damage identification procedure is a multi-level process that encompasses damage detection, damage localization, damage quantification, and residual life prediction \citep{rytter1993vibrational}. By leveraging on sensing, data acquisition and data transfer systems, structural health monitoring (SHM) techniques are aimed at identifying the damage by means of suitable processing and analysis methods \citep{karakostas2023seismic}. Damage detection can be possibly carried out by means of pure data-driven methods, and it is especially important to prevent costly maintenance plans or, even worse, important social, economic and environmental consequences. Early detection of damage is crucial to optimize proactive maintenance interventions before  the onset of  a critical system state. Vibration-based methods are very common in this context because of the availability of low-cost sensing solutions that can be deployed with minimal interference with the construction. Their implementation is based on the premise that the damage can alter the overall dynamic behavior of the structure. The main advantage is that damage can be detected from the measured dynamic response of the structure at the very early stage without the need to be developed to the level that it is visible.

Several vibration-based methods have been developed either in frequency or time domain by assuming that the structure exhibits a linear behavior before and after damage \citep{hou2021review}. The widespread use of mode shapes, natural frequencies and damping ratios for damage detection in linear dynamic systems is supported by well-established output-only techniques for operational modal analysis of structures. In particular, modal curvatures have proven to be especially useful in detecting damages in linear structures \citep[e.g., ][]{pandey1991damage,quaranta2016damage}. Conversely, natural frequencies turn out to be almost insensitive to several damage scenarios \citep[e.g., ][]{doebling1998summary,fan2011vibration} while damping ratios do not always provide consistent results \citep[e.g., ][]{salawu1995bridge,cao2017structural}. Time-domain methods have been also explored for damage detection in linear structures \citep{fassois2007time}, and they usually leverage on state-space \citep[e.g., ][]{dohler2014structural,gres2021subspace} or autoregressive-type models \citep[e.g., ][]{sohn2001damage,lakshmi2017singular}. 

Damage detection in case of nonlinear dynamics has received considerably less attention. Initial attempts were based on the assumption that the undamaged structure behaves linearly while the occurrence of damage causes the transition to a nonlinear response \citep{worden2008review}. However, the hypothesis of initial linear behavior does not hold true for a large class of dynamic systems. Some vibration-based methods have been thus developed for initially nonlinear dynamic systems. For instance, \cite{adams2002nonlinear} have shown that model reduction near bifurcations can be useful to identify certain features that can facilitate damage detection in nonlinear dynamic systems. The extension of the modal-based damage detection approach to nonlinear dynamic systems has been pursued by \cite{lacarbonara2016nonlinear} through the use of nonlinear normal modes. This analytical study proved that the sensitivity of the effective nonlinearity coefficients regulating the nonlinear modal backbones is higher than the sensitivity of the linear frequencies  with respect to damage. Such theoretical evidence was later confirmed in some experimental results reported by \cite{civera2019using} and \cite{carboni2022nonlinear}. \cite{prawin2019damage} addressed the damage detection for nonlinear dynamic systems in the frequency domain using an improved describing function. Among the existing time-domain methods, several proposals deal with the use of Volterra series since such  method provides an attractive way to describe the response of nonlinear dynamic systems. For example, \cite{shiki2017application} and \cite{villani2019damage,villani2019damageExp}  applied the discrete-time Volterra series and a stochastic version of Volterra series, respectively, to detect damage in dynamic systems exhibiting a nonlinear behavior even in the undamaged configuration. The output-only version of the Volterra series proposed by \cite{peng2021nonlinear} for nonlinear structural damage detection employs the structural responses at two different locations in order to identify the kernel function parameters and to evaluate the contribution of the nonlinear components. 

Recent advances in machine learning methods are paving the way towards significant advances in damage detection. In particular, there is great potential in deep learning algorithms for data-driven, vibration-based damage detection \citep{avci2021review}. Autoencoders (AEs) and generative adversarial networks (GANs) appear to be among the most promising deep learning architectures for this task, although their applications are still limited and primarily focused on linear dynamic systems. Both these architectures employ an unsupervised learning paradigm, which obviates the need for extensive labeled datasets for training \citep{qi2020small}. For example, \cite{pathirage2018structural} proposed an AE-based damage detection approach where the natural frequencies and mode shapes of the linear dynamic system are employed  as input data. The AE implemented by \cite{ma2020structural} utilizes the measured dynamic response for detecting damage in a linear system while the application reported by \cite{shang2021vibration} is based on the use of the multi-dimensional cross-correlation functions computed from linear vibrations. \cite{boccagna2023unsupervised} performed damage detection via AE by using univariate statistics estimated from the measured linear dynamic response after performing dimensionality reduction through principal component analysis. The AE-based damage detection approach proposed by \cite{li2023structural} makes use of the power cepstral coefficients extracted from the acceleration response of a linear dynamic system to provide a compact representation of its linear modal properties. Applications of GANs for damage detection are even fewer. The state-of-the-art review by \cite{avci2021review} highlights the lack of studies on the use of GANs for damage detection in civil structures until 2019. Some examples of vibration-based damage detection in linear dynamic systems via GANs appeared only recently \citep{luleci2023cyclegan,luo2023unsupervised}.

Machine learning algorithms applied for data-driven, vibration-based damage detection in nonlinear dynamic systems are infrequent. A seminal work in this context is due to \cite{bornn2010damage}, who reported an experimental application about the use of the autoregressive support vector machine for damage detection in a small-scale laboratory shear-type system subject to random base motion with constant intensity. This system exhibited nonlinear behavior due to impacts and experienced damage from completely loosened bolts in a single column. The use of deep learning techniques has only recently started to emerge. For instance, 
\cite{jin2023early} successfully applied a deep multi-kernel extreme learning machine for fault diagnosis of rotating machineries. \cite{kim2022enhanced} proposed the application of AEs for the identification of reduced-order model while \cite{zhang2023damage} presented a numerical investigation about the use of AEs for damage detection in conventional bearings for seismic isolation. In this case, the damage was simulated through the complete loss of functionality of one or two bearings at the corners of the isolation level, whereas seismic ground motion records were selected in such a way to match a target seismic spectrum. Nonlinear dynamical systems identification through GANs was recently addressed by \cite{yu2024physics}, but no proposals are known to exist about their application in damage detection problems. 

Indeed, output-only, vibration-based damage detection in time-domain of initially nonlinear systems is very challenging. This is because the features of nonlinear dynamic responses can change considerably in time, especially under varying excitation conditions. As a consequence, discerning the effects induced by damage from those caused by the nonlinear response becomes a truly  challenging task \citep{bornn2010damage}. This is a significant concern for output-only vibration-based damage detection by means of machine learning methods that deserves in-depth  investigations. 

This work explores the use of deep learning algorithms for data-driven, vibration-based damage detection in nonlinear dynamic systems. Within this framework, the original contribution here lies in a comprehensive investigation aimed at evaluating the comparative  effectiveness and performance of AEs and GANs. The remaining part of the present work is organized as follows. Sections \ref{sec:autoencoder} and \ref{sec:gan} illustrate the adopted deep learning architectures for damage detection in nonlinear dynamic systems. Section \ref{sec:num} presents the results of an extensive numerical investigation while Section \ref{sec:exp} illustrates an experimental application. The most important outcomes of the present research are finally included in Section \ref{sec:end}.

%%%%%%%%%%%%%%%%%%%%%%%%%%%%%%%%%%
\section{Autoencoder for damage detection}\label{sec:autoencoder}

The autoencoder (AE) originally proposed by \cite{rumelhart1986learning} is an unsupervised algorithm aimed at performing the input reconstruction with the least possible amount of distortion. Its main components are the encoder, a latent feature representation, and the decoder. Since comprehensive state-of-the-art reviews about AEs are available \cite[e.g., ][]{kiranyaz20211d,yang2022autoencoder,li2023comprehensive}, only details related to the adopted implementation for damage detection applications in nonlinear dynamic systems are presented hereafter.

Let $\xbf \in \mathbb{R}^n$ be the normalized time-history response of the nonlinear dynamic system, the encoder is a function ${\bf g} : \mathbb{R}^n \rightarrow \mathbb{R}^m$ such that:
\begin{equation} 
\hbf = {\bf g}(\xbf),
\label{eq:encoder}
\end{equation}
where  $\hbf \in \mathbb{R}^m$ is the latent feature representation of $\xbf$. The decoder is another function ${\bf f} : \mathbb{R}^m \rightarrow \mathbb{R}^n$ such that:
\begin{equation} 
\tilde{\xbf} = {\bf f}(\hbf)={\bf f}({\bf g}(\xbf)),
\label{eq:deencoder}
\end{equation}
where $\tilde{\xbf} \in \mathbb{R}^n$ is the input reconstruction.
In most typical AE architectures, the functions $\bf f$ and $\bf g$ representing the encoding and decoding blocks are artificial neural networks (NNs). In particular, convolutional NNs (CNNs) are mainly employed in AEs because they excel at recognizing the most important signal features and at capturing relevant patterns in the data. 1D CNNs are preferred in the present work because of their minimum computational complexity. It is noted that a normalized input is fed into the AE algorithm so as to reduce the variability among data points from different scales and to speed up the convergence of the networks during training phase. Scaling the signals to a uniform bounded range mitigates the dominance of higher magnitude features over smaller ones, which allows to recognize subtle patterns of damage without bias towards more prominent signal features.  

The components of the implemented CNN-based AE architecture are input layer, convolutional layer, pooling layer, fully connected layer and output layer. The input layer of the CNN aims at importing the normalized 1D time-history response data of the nonlinear dynamic system. Data normalization is accomplished through min-max scaling within the range $[0,1]$. A deep AE is here considered, which means that it involves more than one convolutional layer in the encoding and decoding blocks. Within a convolutional layer, the convolution operator applies on a given input (i.e., the output from the previous layer) using a convolution filter to produce a feature map, whereas a nonlinear activation function is next employed to obtain the output features. The convolutional operation involves sliding a small filter (also known as kernel) and computing dot products between the filter weights and the input values at each location as follows: 
\begin{equation}
{^{\ell+1}}y_i(j) = {^\ell}\Kbf_i * {^\ell}\hat{\xbf}(j) + {^\ell}b_i,
\label{eq:convolution}
\end{equation}
where ${^\ell}\Kbf_i$ and ${^\ell}b_i$ denote the weights and the bias of the $i$th filter in the $\ell$th layer, respectively, ${^\ell}\hat{\xbf}(j)$ is the $j$th receptive field  in the $\ell$th layer  (i.e., the local region representing the portion of the input values that is being processed at a given step of the convolution operation), and ${^{\ell+1}}y_i(j)$ is the input of the $j$th artificial neuron in the frame $i$ of the following $(\ell+1)$th layer. One filter corresponds to one frame in the following layer, and the number of frames is also known as layer depth. It is noted that the same set of learnable parameters (weights) across different receptive fields is utilized. Such weight-sharing allows the network to detect patterns and features regardless of their location. This is because the same filter is applied across the entire input,  reducing the computational effort. Zero padding is applied to both the beginning and the end of the input to preserve the original length of the convolution result. This approach enhances the network's ability to learn features at the boundaries, which is particularly important for short inputs. The stride, defined as the number of positions by which the filter moves across the input during the convolution operation, is chosen to ensure consistent alignment between the input and output across the different layers of the network. 

The implementation of artificial neurons with a nonlinear activation function after the convolution operations is of utmost importance. In fact, without a nonlinear activation function, the output of each convolution would be a linear transformation of the input, which is unfit to capture inherent complex patterns from the data. Conversely, introducing nonlinearity allows the network to learn and represent complex relationships in the input. In the present study, the Leaky ReLU activation function is utilized. It was proposed to address one of the main limitations of the traditional ReLU activation function, which is due to the fact that it sets all negative inputs to $0$, potentially leading to vanishing gradients \citep{glorot2011deep,maas2013rectifier}. The Leaky ReLU activation function mitigates this issue by introducing a small, non-null gradient when the activation function is negative as follows:
\begin{equation}
{^{\ell+1}}a_i(j) = \psi({^{\ell+1}}y_i(j))=
\begin{cases} 
{^{\ell+1}}y_i(j)  & \text{if } {^{\ell+1}}y_i(j) > 0  \\
{^{\ell+1}}\alpha y_i(j)  & \text{otherwise} 
\end{cases},
\label{eq:activation}
\end{equation}
where ${^{\ell+1}}a_i(j)$ is the activation value of the convolution operation result ${^{\ell+1}}y_i(j)$ in Eq. \eqref{eq:convolution} while $\alpha$ is the slope coefficient, which controls the angle of the negative slope in case of negative input values. By using the Leaky ReLU activation function, the network can learn from negative inputs, enhancing its robustness and ability to capture complex patterns. Furthermore, using min-max scaling in combination with Leaky ReLU activation function ensures that the majority of input falls into an active learning range, reducing the risk of inactive neurons.

A pooling layer after the convolutional layer is meant at lowering both the size of the features and the parameters of the network. This, in turn, forces the network to learn only the most important features of the input. Additionally, it reduces training time and mitigates overfitting. The max-pooling layer is employed. Let ${^\ell}q_i(s)$ be the value of the artificial neuron $s$ in the $i$th frame of the $\ell$th layer, the max-pooling transformation applies as follows:
\begin{equation}  
{^{\ell+1}}p_i(j) = \max_{(j-1)W+1 \le s \le jW}{{^\ell}q_i(s)},
\label{eq:maxpooling}
\end{equation}
where ${^{\ell+1}}p_i(j)$ is the corresponding value of the artificial neuron in the $(\ell + 1)$th layer of the pooling operation and $W$ is the pooling width.

By adjusting padding and stride, convolutional layers for the encoder are designed to downscale the input together with pooling layers, while convolutional layers in the decoder are intended to upscale the input (through transposed convolution operations, and pooling layers do not take place in this case). Convolutional layers also serve as output layer of the AE. Conversely, fully connected layers are implemented in the latent feature representation to effectively lower the dimensionality of the input, as this task would be cumbersome with convolutional layers. Controlling the size of the latent representation is important because it plays an important role in determining how the machine learning model performs on the assigned data. A fully connected layer, also known as dense layer, is a standard layer where each artificial neuron is connected to every artificial neuron in the previous layer, similar to a traditional NN \citep[e.g., ][]{quaranta2020review}. Before the fully connected layer applies, the output from the last pooling layer in the encoding block is flattened. 

The AE is required to reconstruct the input time-history response of the nonlinear dynamic system  well enough while producing a meaningful latent representation in such a way to avoid the overfitting phenomenon. This is accomplished through proper training of the AE, which basically consists of finding the optimal set of hyperameters $\thetabf$ ruling the functions ${\bf f}$ and ${\bf g}$ that represent the encoding and decoding blocks. To this end, two main strategies leveraging on parsimony are adopted \citep{kutz2022parsimony} and are based on creating a bottleneck and adding a regularization term. A bottleneck involves reducing the dimensionality $m$ of the latent features compared to that of the data $n$. Including a regularization term in the training process also prevents the model from fitting the noise in the training data too closely. In the present work, the $L_2$ technique is implemented, and thus the AE training seeks the solution of the following optimization problem: 
\begin{equation} 
\min_{ \thetabf }{ \left\{ \textsf{E} \left[ \Delta \left(\xbf,f(g(\xbf);\thetabf) \right) \right] + \lambda \|\thetabf\|_2^2 \right\} },
\label{eq:training}
\end{equation} 
where the first term is a loss function that quantifies how input and output of the AE differ each other, on average, across all $\xbf$ ($\textsf{E}$ represents the expectation operator), while $\lambda$ and $\|\thetabf\|_2$ in the second term are the regularization coefficient and the $L_2$-norm of the hyperameters $\thetabf$ involved in the functions ${\bf f}$ and ${\bf g}$, respectively. The adopted loss function in the present study is the mean absolute error (MAE). The set of tuned hyperparameters $\thetabf$ includes kernel size, filter numbers, batch size (i.e., number of input values processed simultaneously during one training step), and latent filters. The values of slope coefficient $\alpha$ and regularization coefficient $\lambda$ are instead established before training (they are set to $0.2$ and in the range $10^{-4}$ to $10^{-8}$, respectively). 

The adaptive moment estimation (ADAM) optimizer is adopted to solve Eq. \eqref{eq:training}. It is a gradient-based technique \citep{kingmaAdamMethodStochastic2017} that combines the advantages of the adaptive gradient algorithm and root mean square propagation. ADAM computes adaptive learning rates for each hyperparameter based on estimates of the first and second moments of the gradient. The update rules in ADAM are the following:
\begin{subequations} 
\begin{equation} 
\mbf_\tau = \beta_1 \mbf_{\tau-1} + (1 - \beta_1) \nabla_\tau,
\end{equation}
\begin{equation}
\vbf_\tau = \beta_2 \vbf_{\tau-1} + (1 - \beta_2) \nabla_\tau^2,
\end{equation}
\begin{equation}
\hat{\mbf}_\tau = \frac{\mbf_\tau}{1 - \beta_1^\tau},
\end{equation}
\begin{equation}
\hat{\vbf}_\tau  = \frac{\vbf_\tau}{1 - \beta_2^\tau},
\end{equation}
\begin{equation}
\thetabf_{\tau+1} = \thetabf_\tau - \frac{\eta}{\sqrt{\hat{\vbf}_\tau} + \epsilon} \hat{\mbf}_\tau,
\end{equation}
\label{eq:adam}
\end{subequations} 
where $\mbf_\tau$ and $\vbf_\tau$ are the updated first (i.e., mean) and second (i.e., uncentered variance) moments of the gradients, respectively, $\beta_1$ and $\beta_2$ are the exponential decay rates for these moment estimates (set to 0.9 and 0.999, respectively), $\nabla_\tau$ is the objective function gradient at the iteration $\tau$, $\eta$ is the learning rate (in the range $10^{-2}$ to $10^{-4}$), and $\epsilon$ is a small scalar added to improve numerical stability (set to $10^{-8}$), while $\thetabf_{\tau+1}$ represents the updated hyperparameters for the next iteration (aka epoch), or the outcome of the optimization if the convergence criterion is fulfilled. To further strengthen the network against overfitting phenomenon, early stopping applies by setting a low value of the patience (i.e., the number of epochs the training process continues without improvement) between 4 and 6.    

\begin{figure}[h!]
\centering
\includegraphics[width=.7\linewidth]{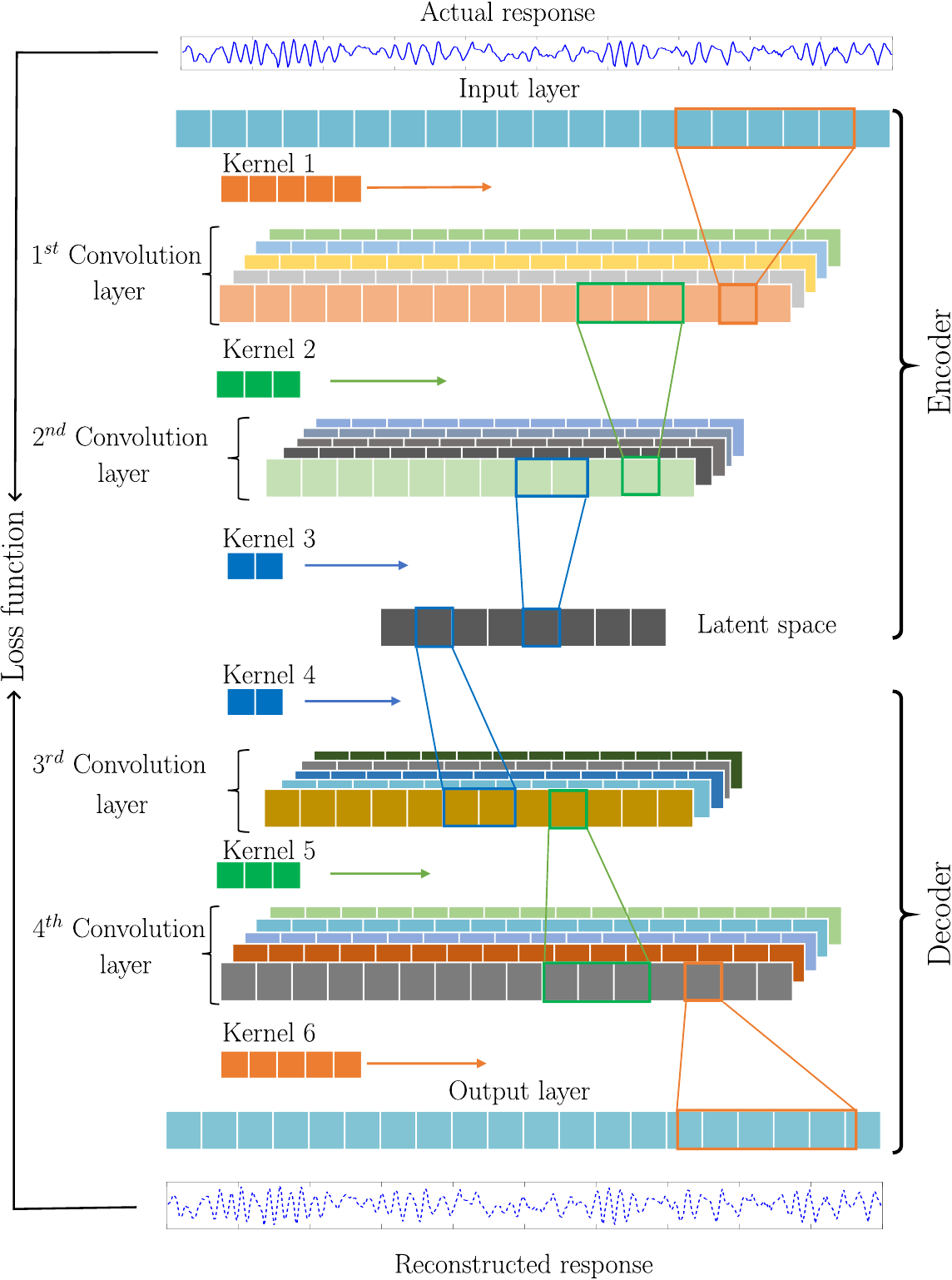}
\caption{Implemented AE architecture based on 1D CNNs for damage detection in nonlinear dynamic systems.}
\label{fig:AElayout}
\end{figure}

The suitability of the AE for data-driven damage detection relies on the evidence that, once properly trained over time series data collected from an undamaged dynamic system, it will turn into a large loss function value when attempting to reconstruct the response of a damaged one. Figure \ref{fig:AElayout} illustrates the implemented AE architecture based on 1D CNNs for damage detection in nonlinear dynamic systems. The corresponding algorithm was developed using TensorFlow \citep{tensorflow2015-whitepaper}. 

%%%%%%%%%%%%%%%%%%%%%%%%%%%%%%%%%%
\section{Generative adversarial  network for damage detection}\label{sec:gan}
A generative adversarial network (GAN) is an unsupervised algorithm originally proposed by \cite{goodfellow2014generative} that consists of two models called generator and discriminator. The generator tries to create realistic data,  while the discriminator aims to differentiate between real data and the fake data created by the generator. The details concerning the specific implementation of the GAN for damage detection applications in nonlinear dynamic systems will be presented hereafter while more general information can be found elsewhere \cite[e.g., ][]{gui2021review,creswell2018generative,xia2022gan}.

On the one hand, the generator can be represented as a differentiable function $G(\zbf ; \thetabf_G)$ where $\zbf$ is a random noise vector from a distribution $\varphi_{\zbf}(\zbf)$ and $\thetabf_G$ is a set of hyperparameters. It produces synthetic data samples, which should resemble real data samples. On the other hand, the discriminator is another differentiable function $D(\xbf ; \thetabf_D)$ where $\xbf$ is a data sample from a distribution $\varphi_{data}(\xbf)$ and  $\thetabf_D$ is a set of hyperparameters. It produces a scalar output representing the probability that $\xbf$ comes from the real data distribution rather than the generator's distribution. Data normalization through min-max scaling is also performed in GAN to enhance the performance of the algorithm as well as the efficiency of the training phase.

The training of GAN is a non-cooperative game. Hence, GAN optimization is attained at the Nash equilibrium point \citep{ratliff2013characterization} corresponding to a stable state where neither the generator, nor the discriminator obtains a beneficial effect (i.e., gain) by deviating from the current configuration. At this equilibrium point, the generator produces data that are so realistic that the discriminator cannot distinguish whether they are real or fake, and the discriminator cannot improve its accuracy by further adjustments. The training of discriminator $D$ and generator $G$ in the GAN requires the solution of the following optimization problems:
\begin{equation}
\max_{\thetabf_D}\{ \textsf{E}_{\xbf \sim \varphi_{data}(\xbf)}[\log D(\xbf;\thetabf_D)] + \textsf{E}_{\zbf \sim \varphi_{\xbf}(\zbf)}[\log{(1-D(G(\zbf|\thetabf_G);\thetabf_D))}] \},
 \label{eq:discriminatorTraining}
\end{equation}
\begin{equation}
\min_{\thetabf_G}\{ \textsf{E}_{\zbf \sim \varphi_{\zbf}(\zbf)}[\log{(1-D(G(\zbf;\thetabf_G)|\thetabf_D))}] \}.
 \label{eq:generatorTraining}
\end{equation}
Altogether, Eqs. \eqref{eq:discriminatorTraining}-\eqref{eq:generatorTraining} are equivalent to the following min-max optimization problem:
 \begin{equation}
\min_{\thetabf_G}{ \max_{\thetabf_D}\{ \textsf{E}_{\xbf \sim \varphi_{data}(\xbf)}[\log D(\xbf;\thetabf_D)] + \textsf{E}_{\zbf \sim \varphi_{\xbf}(\zbf)}[\log{(1-D(G(\zbf;\thetabf_G);\thetabf_D))}] \} },
\label{eq:minmaxGAN}
\end{equation}
wherein the objective function is related to the Jensen-Shannon divergence between  $\varphi_{data}(\xbf)$ and $\varphi_{\zbf}(\zbf)$ \citep{gui2021review} ($\textsf{E}$ represents the expectation concerning the distribution appointed in the subscript). The GAN version based on the optimization problem in Eq. \eqref{eq:minmaxGAN} is commonly referred to as vanilla GAN. 

A deep GAN is here implemented by stacking multiple 1D CNNs. In particular, 1D convolutional layers according to Eqs. \eqref{eq:convolution}-\eqref{eq:activation} are used for both generator and discriminator (wherein transposed convolution operations are employed for the generator). In this regard, a significant novelty deals with the output layer of the discriminator, which implements a Sigmoid activation function as follows: 
\begin{equation}
{^{\ell+1}}a_i(j) = \psi({^{\ell+1}}y_i(j)) = \dfrac{{^{\ell+1}}y_i(j)}{1+\text{e}^{{^{\ell+1}}y_i(j)}}.
\end{equation}
The hyperparameters are tuned alternatively solving Eqs. \eqref{eq:discriminatorTraining}-\eqref{eq:generatorTraining} throughout the training stage by means of ADAM update rules in Eq. \eqref{eq:adam}. During training, the hyperparameters of one model are updated, while the hyperparameters of the other are fixed. The set of tuned hyperparameters for GAN includes number filters, kernel size, stride length, and depth of the neural network (these hyperparameters are different for the generator and discriminator). The training process for GAN runs until a number of epochs between $1,000$ and $2,000$ is achieved (no early stopping applies in this case). The regularization of GAN discriminator is accomplished via dropout. Accordingly, throughout the training, an artificial neuron is temporarily dropped at each iteration (i.e., all the inputs and outputs to this artificial neuron are disabled at the current iteration)  with an assigned probability. The dropped-out artificial neurons are resampled with the same probability at every training step, so a dropped out neuron at one step can be active at the next one. 

There are no known proposals regarding the use of GAN in damage detection problems. Nonetheless, the suitability of GAN for data-driven damage detection stems on the fact that, upon proper training with time series data collected from an undamaged system, the discriminator will be no longer able to differentiate between actual undamaged data and synthetic data produced by the generator. Conversely,  the discriminator will label the data from a damaged system as fake because they do no longer resemble those of the undamaged system. In this regard, it is interesting to note that, by implementing the Sigmoid activation function at the output layer of the discriminator, a quantitative assessment of the dynamic system state is obtained from the discriminator. In fact, numerical values of the discriminator output close to $1$ are associated with an undamaged state while lower values approaching to 0 highlight a deviation from the (trained) undamaged condition, thereby pointing out the presence of the damage. Hence, the discriminator output can be exploited to compute a damage index.

\begin{figure}[h!]
\centering
\includegraphics[width=\linewidth]{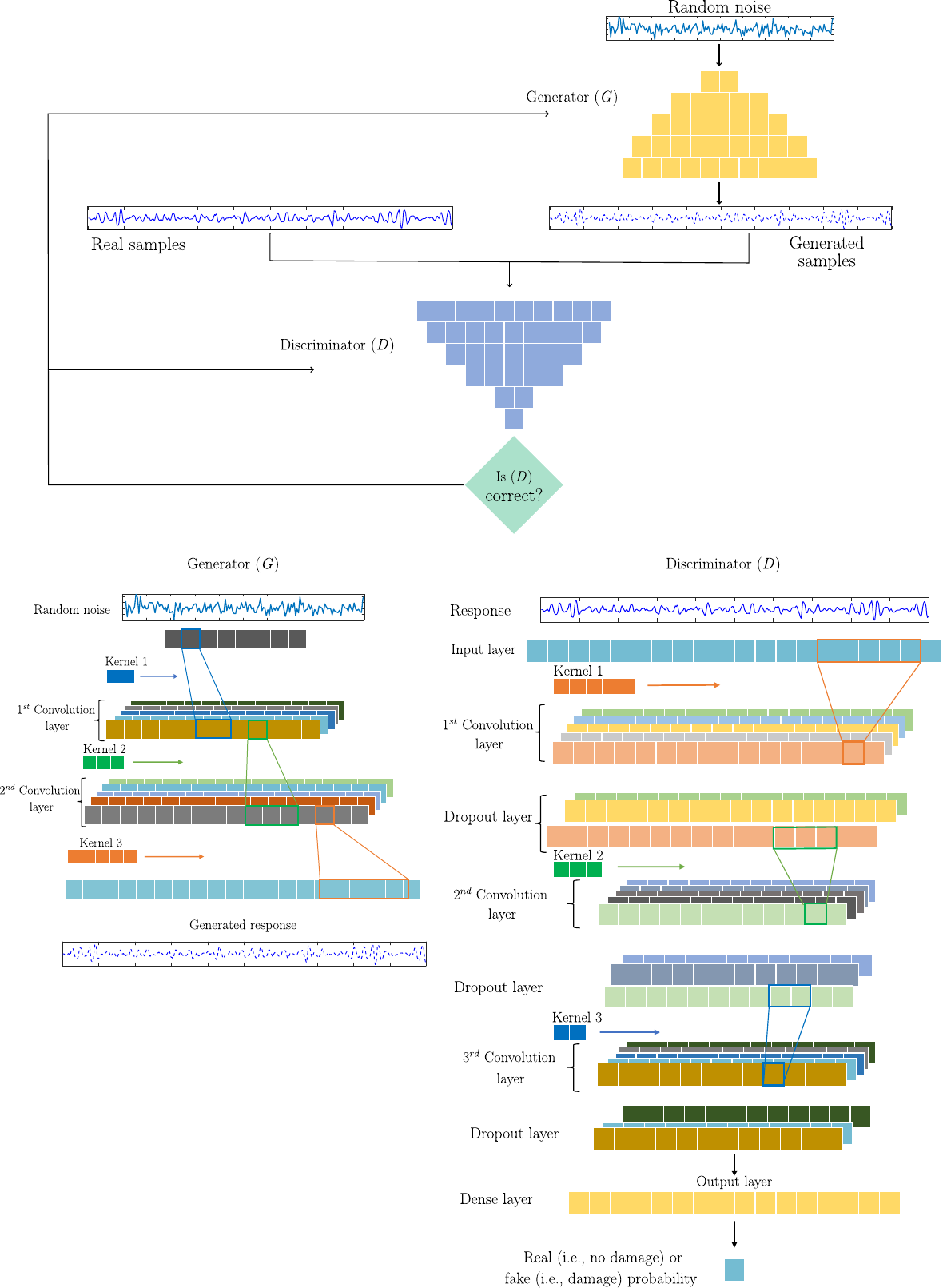}
\caption{Implemented GAN architecture based on 1D CNNs for damage detection in nonlinear dynamic systems.}
\label{fig:GAN}
\end{figure}

This vanilla GAN architecture based on 1D CNNs for damage detection in nonlinear dynamic systems has been implemented using TensorFlow \citep{tensorflow2015-whitepaper}. Figure \ref{fig:GAN} shows pictorially the architecture of the GAN. 

%%%%%%%%%%%%%%%%%%%%%%%%%%%%%%%%%%
\section{Numerical investigation}\label{sec:num}

Numerical simulations are initially performed to evaluate the effectiveness of both AE and GAN in detecting damage within nonlinear dynamic systems from their time-domain response only. To replicate the most common and challenging scenario, the nonlinear dynamic systems under consideration are subjected to random vibrations of varying intensity, and the response is contaminated by white Gaussian noise with a noise level of 10\%. 

The database for training and validating both deep learning architectures include samples of undamaged system displacements over $2$ s time windows, assuming a sampling rate of 250 Hz.  A total of 2,000 time series was collected for such tasks: 80\% of the generated time-history samples is utilized for training the algorithms (i.e., training set), while the remaining 20\% is allocated for their validation (i.e., validation set). The validation dataset was used to check if the training of the network converged properly, avoiding overfitting or underfitting. The full database thus consists of about 1 h of recording of the undamaged system response under different magnitudes of the input random excitation. It is highlighted that neither the applied dynamic loading conditions, nor the dynamic system characteristics, are ultimately involved in the implementation of the considered deep learning architectures. This means that both AE and GAN are meant to perform output-only damage detection of unknown dynamic systems.   

The selected configuration of AE and GAN for each application is first established through preliminary simulations, and then its trainable parameters are found. Once both deep learning algorithms have been trained and validated using the samples of undamaged system response, they are employed to detect damage. To this end, 2,000 time series of noisy system displacements, each of  2 s duration, were generated for constant damage levels ranging from 5\% to 30\%. Therefore, both AE and GAN are required to detect damage, if any, from about 1 h of recording of the system response. While low damage levels are considered to test the capability of detecting the onset of damage, the results for increasing damage levels are examined to inspect the proper working of the implemented deep learning schemes.     

\subsection{Single-degree-of-freedom system with cubic nonlinearity}
A large variety of dynamic systems can exhibit a nonlinear behavior for high levels of excitation due to geometric  or material nonlinear effects. The actual response of such systems can be effectively simulated by means of a restoring force that is dependent upon the cube of the displacement, known as a Duffing-type system. Hence, this class of nonlinear dynamic systems provides a relevant benchmark for evaluating the effectiveness of damage detection techniques in practical applications. The equation of motion for a single-degree-of-freedom (1-DOF) system with cubic nonlinearity and linear viscous damping is the following:
\begin{equation}
M \ddot{x} + C \dot{x} + K_1 x + K_3 x^3 = F(t), 
\label{eq:duffing}
\end{equation}
where $x(t)$ represent the displacement of the system at time $t$ (overdots denote differentiation with respect to time $t$), whereas $M$ is the mass and $C$ is the viscous damping coefficient. Moreover, $K_1$ and $K_3$ are the linear and nonlinear stiffness terms, respectively, while $F$ is the external  time-varying  load. The numerical values of the system parameters in this numerical application are set to: $M=412$ kg, $C= 405.95$ kg/s (corresponding to a damping ratio of 1\%), $K_1=1$ kN/mm, $K_3=0.001$ kN/mm$^3$. It is assumed that the  presence of damage corresponds to a reduction of both the linear and nonlinear stiffness by the same amount, without effects on mass and damping. 

\begin{figure}[h!]
\centering
\includegraphics[width=\linewidth]{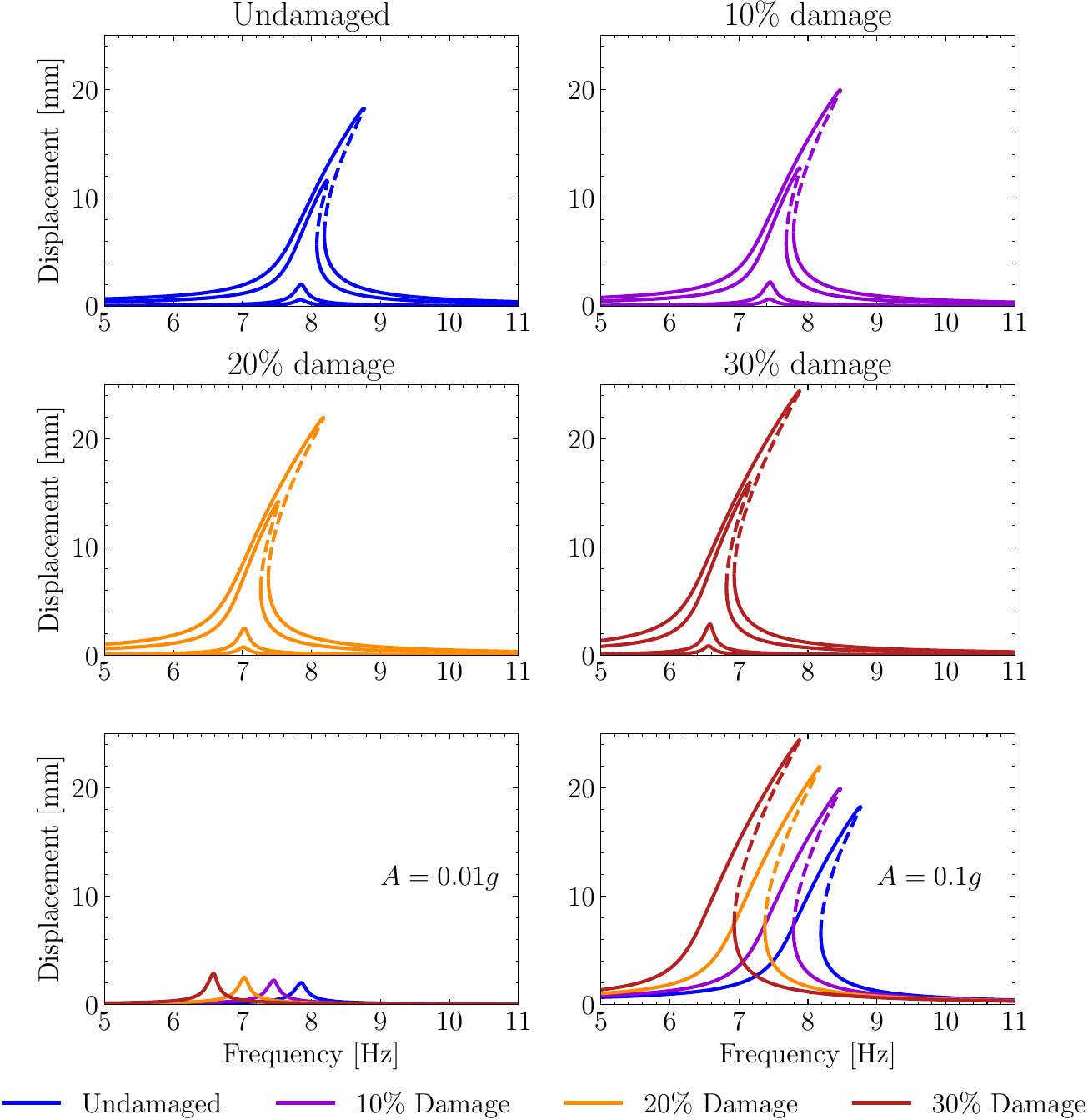}
\caption{Frequency response curves (FRCs) of the 1-DOF system for different excitation amplitudes and damage levels.}
\label{fig:1DOFcubic_FRC}
\end{figure}

The dynamics of such system are first investigated by computing its frequency response curves (FRCs) considering a harmonic  base excitation such that $F(t)=-A M \cos\Omega t$, where $A$ is the peak acceleration value, $\Omega$ is the excitation frequency. Results for the undamaged and damaged system are plotted in Fig. \ref{fig:1DOFcubic_FRC} assuming $A$ between $0.003g$ and $0.1g$. They show that the considered dynamic system exhibits a linear behavior for $A$ less than or equal to $0.01$ g while a nonlinear behavior is apparent otherwise. In particular, a moderate nonlinear response is observed for $A$ between $0.01g$ and $0.06g$, whereas a stronger nonlinear behavior occurs for higher values of the input excitation. For the undamaged system, if $A$ is less than or equal to $0.01g$, then the resonant frequency is 7.8 Hz, while it shifts to $8.8$ Hz (increase by 12.8\%) when $A$ is $0.1g$. 

Figure \ref{fig:1DOFcubic_FRC} is also useful to explain why time-domain, output-only damage detection in nonlinear dynamic systems subject to varying dynamic excitation is so challenging as pointed out by \cite{bornn2010damage}. In fact, Fig. \ref{fig:1DOFcubic_FRC} shows  that the shift of the resonant frequency due to the hardening-type nonlinear response can mislead the fault recognition, thus hindering an accurate early damage detection. 

The effectiveness of both AE and GAN is investigated through numerical simulations by solving Eq. \eqref{eq:duffing} under random excitation such that $F=-\ddot{x}_bM $, where $\ddot{x}_b$ is a white noise. The time-histories displacement response of the undamaged and damaged system are computed once time-history samples of $\ddot{x}_b$ are generated and the corresponding peak values are scaled to a uniform random value between $0.01g$ and $0.1g$ so as to focus on the nonlinear response only, in agreement with Fig. \ref{fig:1DOFcubic_FRC}. Details about the configuration of both AE and GAN are provided in Tab. \ref{table:1DOFcubic}.

\begin{table}[ht]
\centering
\caption{Configuration of the deep learning architectures for damage detection in the 1-DOF system.}
\label{table:1DOFcubic}
\begin{tabular}{@{}llccccc@{}}
\toprule
\textbf{Type} & \textbf{Component} & \textbf{\makecell{Convolution\\layers}} & \textbf{\makecell{Trainable\\parameters}} & \textbf{\makecell{Kernel\\size}} & \textbf{\makecell{Number\\of filters}} & \textbf{Strides} \\ \midrule
\multirow{2}{*}{AE} & Encoder & 3 & 3,767 & (100,50,25) & (10,5,2) & (1,1,1)  \\
                    & Decoder & 3 & 66,056 &(25,50,100)  & (5,10,1)  & (2,2,1) \\
 \midrule
\multirow{2}{*}{GAN}         & Generator & 3 & 16,018 & (2,3,6) & (32,32,1) & (2,1,2) \\
               & Discriminator & 3 &   8,117,729& (6,6,6) &(64,64,64) & (2,2,2) \\ \bottomrule
\end{tabular}
\end{table}

\begin{figure}[h!]
\centering
\includegraphics[width=1\linewidth]{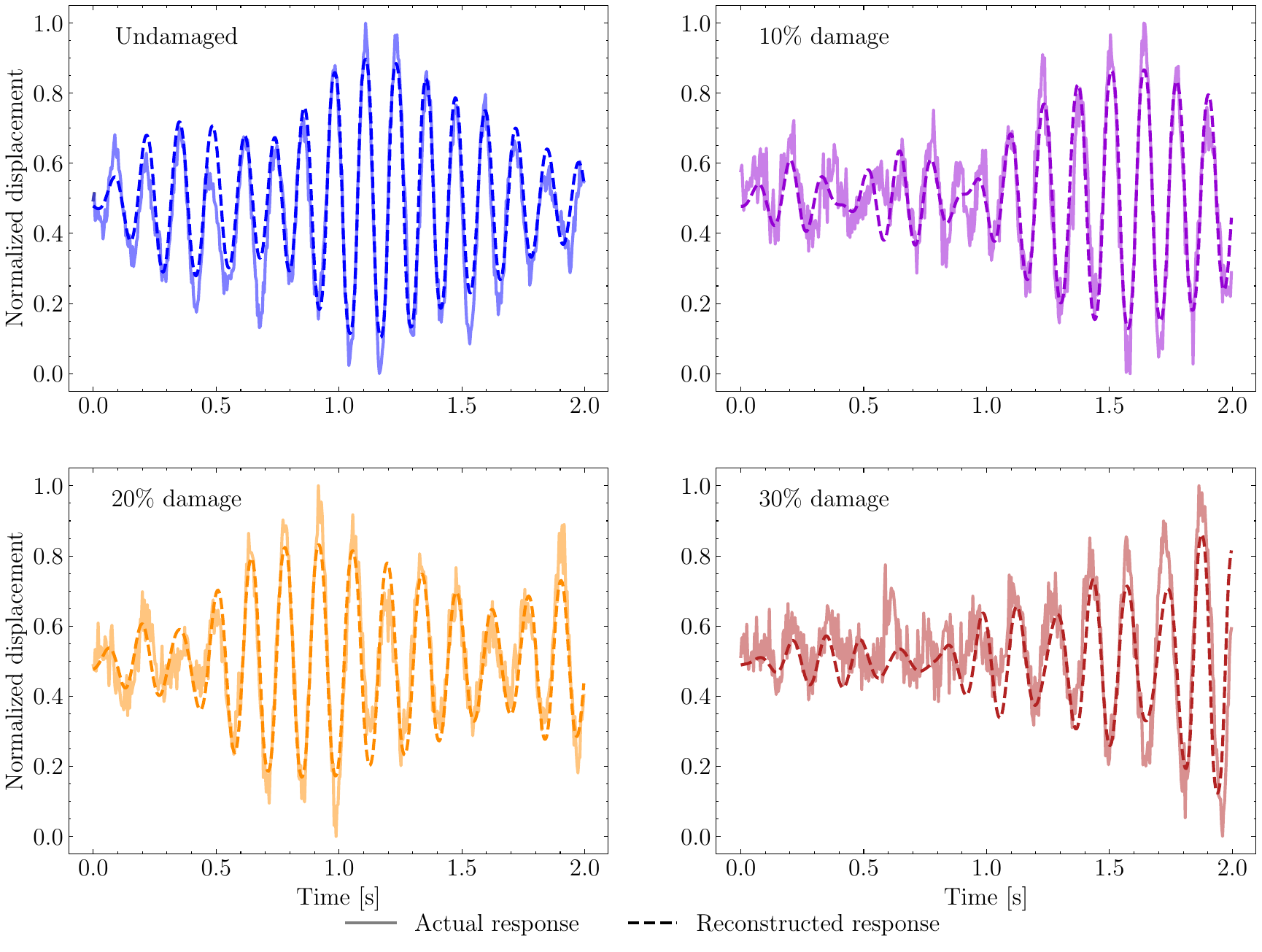}
\caption{Reconstruction of the 1-DOF system response through AE for different damage levels.}
\label{fig:1DOFcubic_ae}
\end{figure}

\begin{figure}[h!]
\centering
\includegraphics[width=1\linewidth]{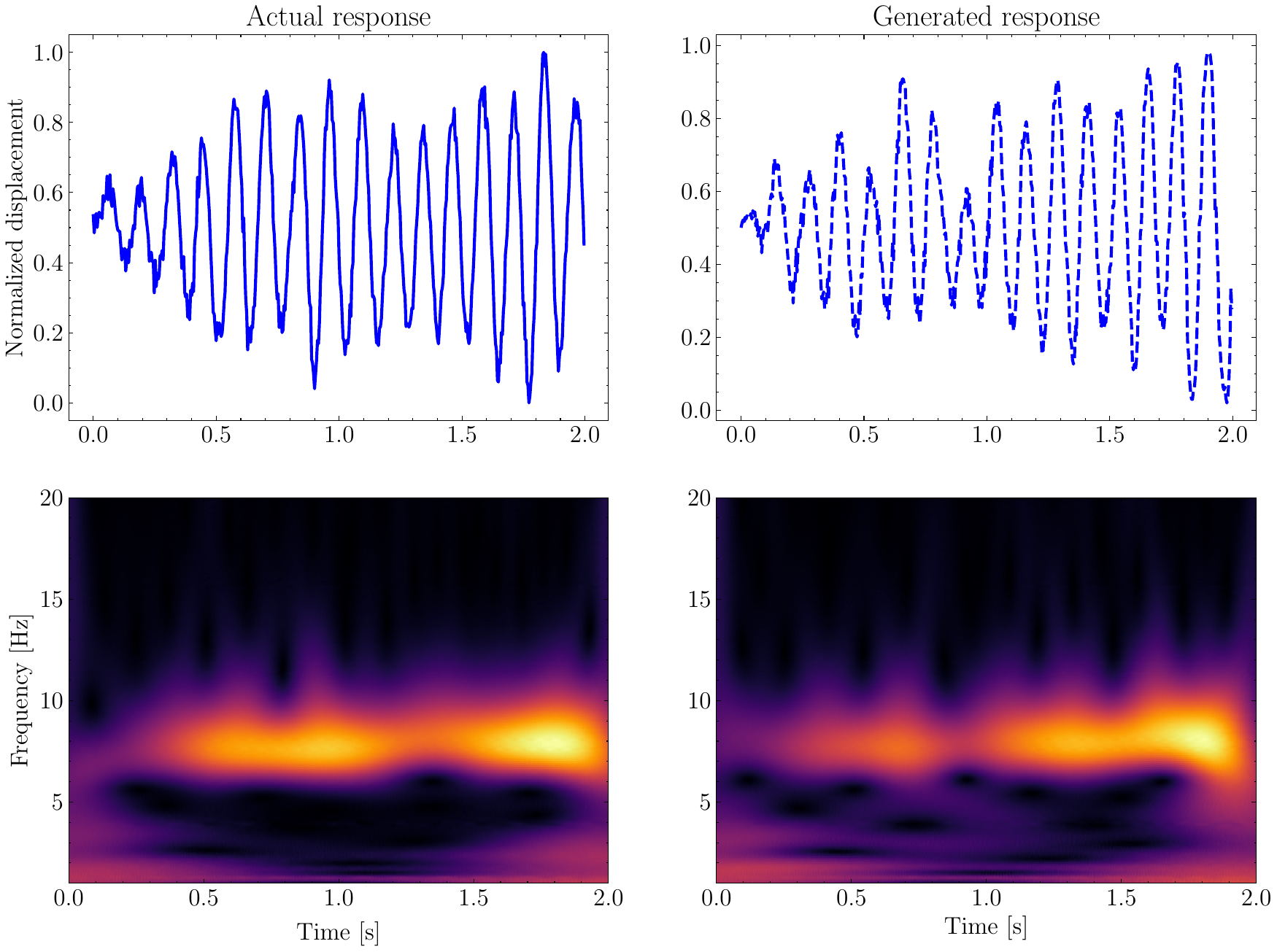}
\caption{Time-history and corresponding time-frequency representation via wavelet transform of the undamaged 1-DOF system response: actual data and results carried out from the GAN generation block.}
\label{fig:1DOFcubic_gan}
\end{figure}

The representative results reported in Fig. \ref{fig:1DOFcubic_ae} confirm that the trained AE is able to reconstruct the time-history displacement of the undamaged nonlinear system (the system response under the highest excitation is plotted herein). Conversely, it is no longer able to achieve similar accuracy levels when it attempts to reconstruct the displacement time-history response of the damaged system. This mismatch between input and output of the AE highlights a deviation from the conditions of the training process, and thus points to the occurrence of the damage. The greater the damage level, the larger the difference between the input and output of the AE. The GAN operates differently, requiring the generator to accurately mimic the undamaged system response. This maximizes the damage sensitivity of the discriminator throughout the training process. In this regard, a representative pair of actual and generated displacement time-histories are plotted in Fig. \ref{fig:1DOFcubic_gan} (the dominant frequency herein is about 8.8 Hz, which corresponds to the highest excitation level according to Fig. \ref{fig:1DOFcubic_FRC}). The high similarity between them in the time-frequency domain demonstrates the fidelity of the GAN in simulating the dynamics of the undamaged nonlinear system. This, in turn, implies that the discriminator has been successfully trained and will be thus able to recognize accurately a deviation from the training conditions, such as the occurrence of damage. 

\begin{figure}[h!]
\centering
\includegraphics[width=1\linewidth]{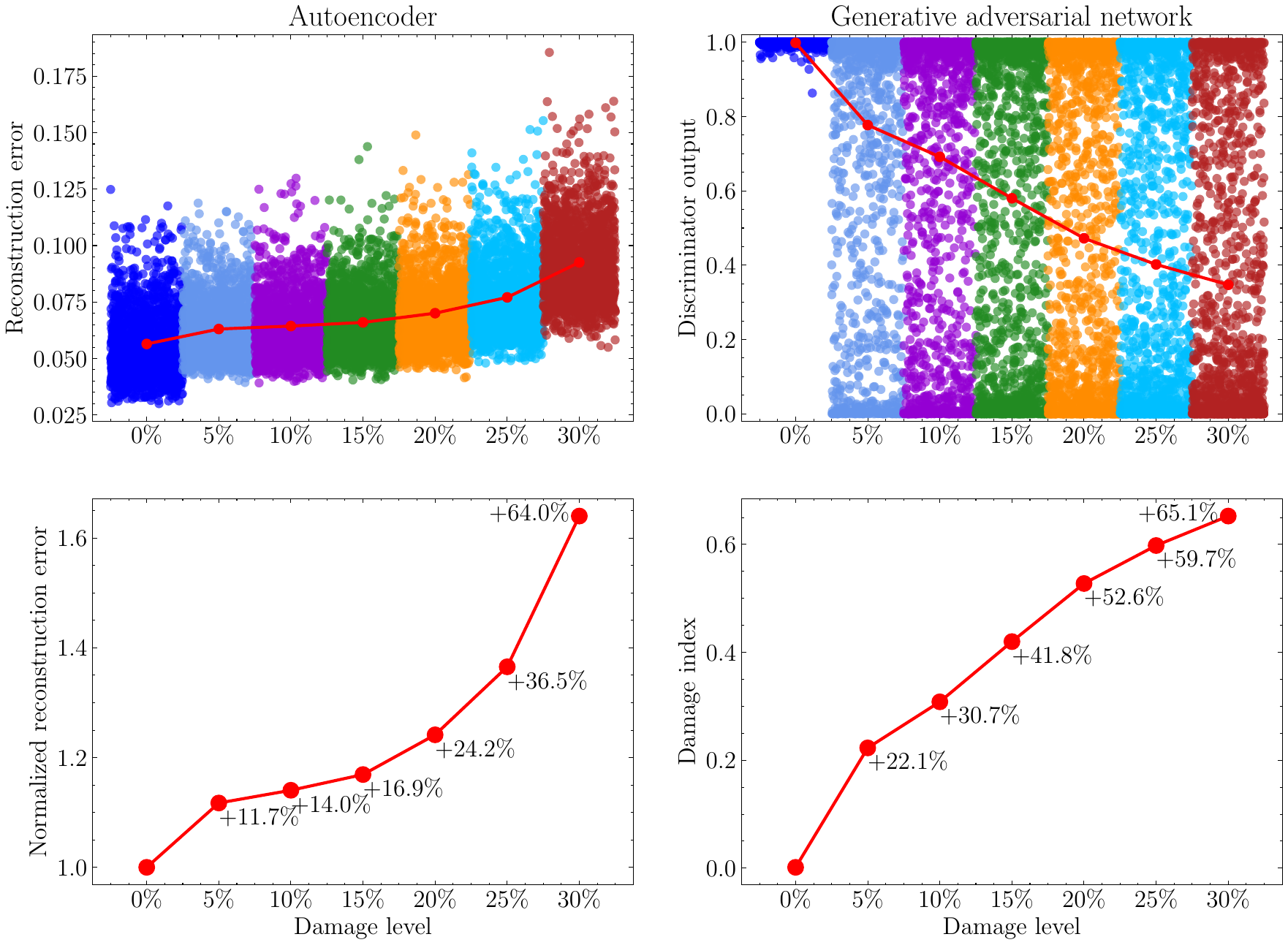}
\caption{Damage sensitivity of the two deep learning architectures for the 1-DOF system: reconstruction loss values obtained by AE and relative variation of its normalized average value, together with the discriminator output values obtained by GAN and relative variation of the associated average damage index.}
\label{fig:1DOFcubic_damage}
\end{figure}

Figure \ref{fig:1DOFcubic_damage} compares the performance of both deep learning architectures in detecting damage. Each dot in the scatter plots represents the reconstruction loss (in terms of MAE) of the AE and the discriminator output of the GAN for each time series sample given the damage level, while the trend lines connect the corresponding average values. Since the discriminator output is bounded between 0 and 1, a damage index is defined as the complement of the discriminator output.  Figure \ref{fig:1DOFcubic_damage} shows that reconstruction loss values shift upwards (i.e., the reconstruction loss increases) as the damage severity grows. It can also be inferred from Fig. \ref{fig:1DOFcubic_damage}  that the discriminator outputs are close to 1 when there is no damage, while a growing number of outcomes approach 0 as the damage level increases. Therefore, the relative variation of the normalized reconstruction loss and that of the damage index based on the discriminator loss properly correlate with the damage level. Notably, a significant deviation from the reference value corresponding to the undamaged configuration already occurs at the lowest level of damage (i.e., relative variation larger than 10\% for the lowest damage severity), which demonstrates a satisfactory sensitivity of both deep learning architectures. 

\subsection{Two-degree-of-freedom system with cubic nonlinearity}
\begin{figure}[h!]
\centering
\includegraphics[width=\linewidth]{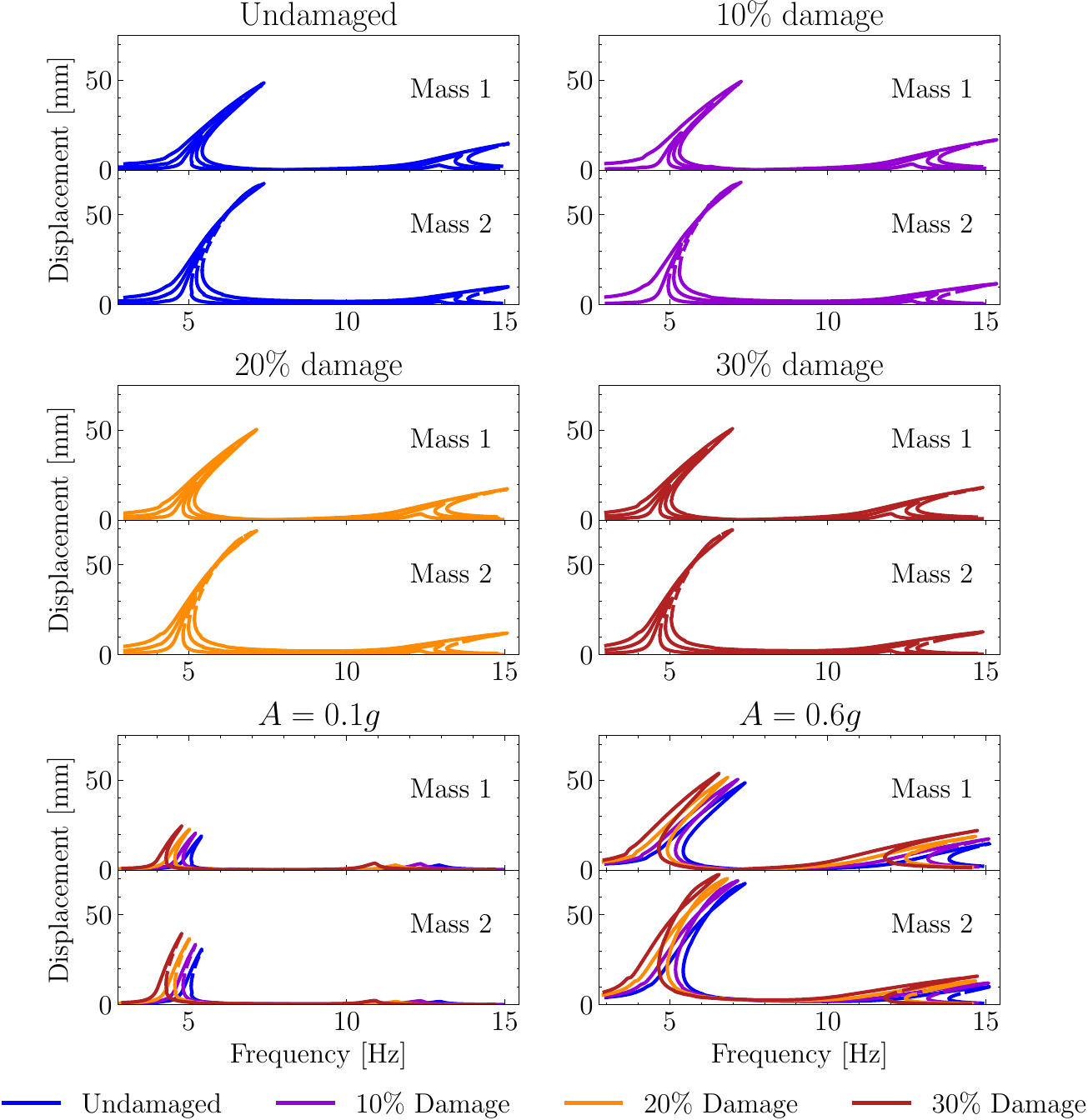}
\caption{Frequency response curves (FRCs) of the 2-DOF system for different excitation  amplitudes and damage levels.}
\label{fig:2DOFcubic_FRC}
\end{figure}

Damage detection in a two-degree-of-freedom (2-DOF) nonlinear dynamic system with cubic nonlinearity and linear viscous damping is considered next. The corresponding equations of motion are the following:
\begin{subequations}
\begin{equation}
\begin{aligned}
M_1\ddot{x}_1 &+ C_1\dot{x}_1 + K_{11} x_1 + K_{13}x_1^3\\ &- C_2\left(\dot{x}_2-\dot{x}_1\right)-K_{21}\left(x_2-x_1\right)-K_{23}\left(x_2-x_1\right)^3 = F_1(t), 
\end{aligned}
\end{equation}
\begin{equation}
M_2\ddot{x}_2 + C_2\left(\dot{x}_2-\dot{x}_1\right)+K_{21}\left(x_2-x_1\right)+K_{23}\left(x_2-x_1\right)^3 = F_2(t),
\end{equation}
\label{eq:2DOFduffing}
\end{subequations}
where $x_1(t)$ and $x_2(t)$ are the displacements of the two oscillators, $M_1$ and $M_2$ are the masses, $K_{11}$ and $K_{21}$ are the linear stiffness coefficients, $C_1$ and $C_2$ are the viscous damping coefficients, $K_{13}$ and $K_{23}$ are the cubic stiffness coefficients, $F_1$ and $F_2$ are the excitation forces.  The numerical values of the system parameters in this numerical application are set to: $M_1=M_2=400$ kg, $C_1=C_2=$ 400 kg/s (corresponding to a damping ratio of 1\%), $K_{11}=K_{21}=1$ kN/mm, $K_{13}=K_{23}=0.001$ kN/mm$^3$. It is assumed that linear and nonlinear stiffness coefficients get reduced by the same amount in both oscillators as a result  of damage, while damage does not affect their masses and damping coefficients. The FRCs of the undamaged and damaged 2-DOF nonlinear system are shown in Fig. \ref{fig:2DOFcubic_FRC} for $F_1=-A M \cos\Omega t$ and $F_2=0$, with $A$ between $0.1g$ and $0.6g$.

A random excitation is employed to detect damage assuming $F_1=-\ddot{x}_bM_1$ and $F_2=0$. The displacement time-histories of both the undamaged and damaged systems are thus calculated through Eq. \eqref{eq:2DOFduffing} by generating independent time-history samples of the white noise $\ddot{x}_b$ and adjusting their peaks to match a uniformly random value between $0.1g$ and $0.6g$ in order to restrict the analysis to a mid-large nonlinear response according to Fig. \ref{fig:2DOFcubic_FRC}. The settings adopted to implement the deep learning architectures are listed in Tab. \ref{table:2DOFcubic}.

\begin{table}[ht]
\centering
\caption{Configuration of the deep learning architectures for damage detection in the 2-DOF system.}
\label{table:2DOFcubic}
\begin{tabular}{@{}llccccc@{}}
\toprule
\textbf{Type} & \textbf{Component} & \textbf{\makecell{Convolution\\layers}} & \textbf{\makecell{Trainable\\parameters}} & \textbf{\makecell{Kernel\\size}} & \textbf{\makecell{Number\\of filters}} & \textbf{Strides} \\\midrule
\multirow{2}{*}{AE} & Encoder &3  & 150,175 & (50,25,12) &(100,50,25)  & (1,1,1) \\
            & Decoder & 4 & 2,142,902 &(12,25,25,50)    &(50,100,100,2)  &  (2,2,1,1)\\
\midrule
\multirow{2}{*}{GAN}         & Generator &3  & 38,151 & (6,6,4) &(64,32,2)  &(2,1,2)  \\
            & Discriminator & 3 & 4,050,770 & (6,6,6) &(32,32,32)  &  (2,2,2)\\ \bottomrule
\end{tabular}
\end{table}

\begin{figure}[h!]
\centering
\includegraphics[width=1\linewidth]{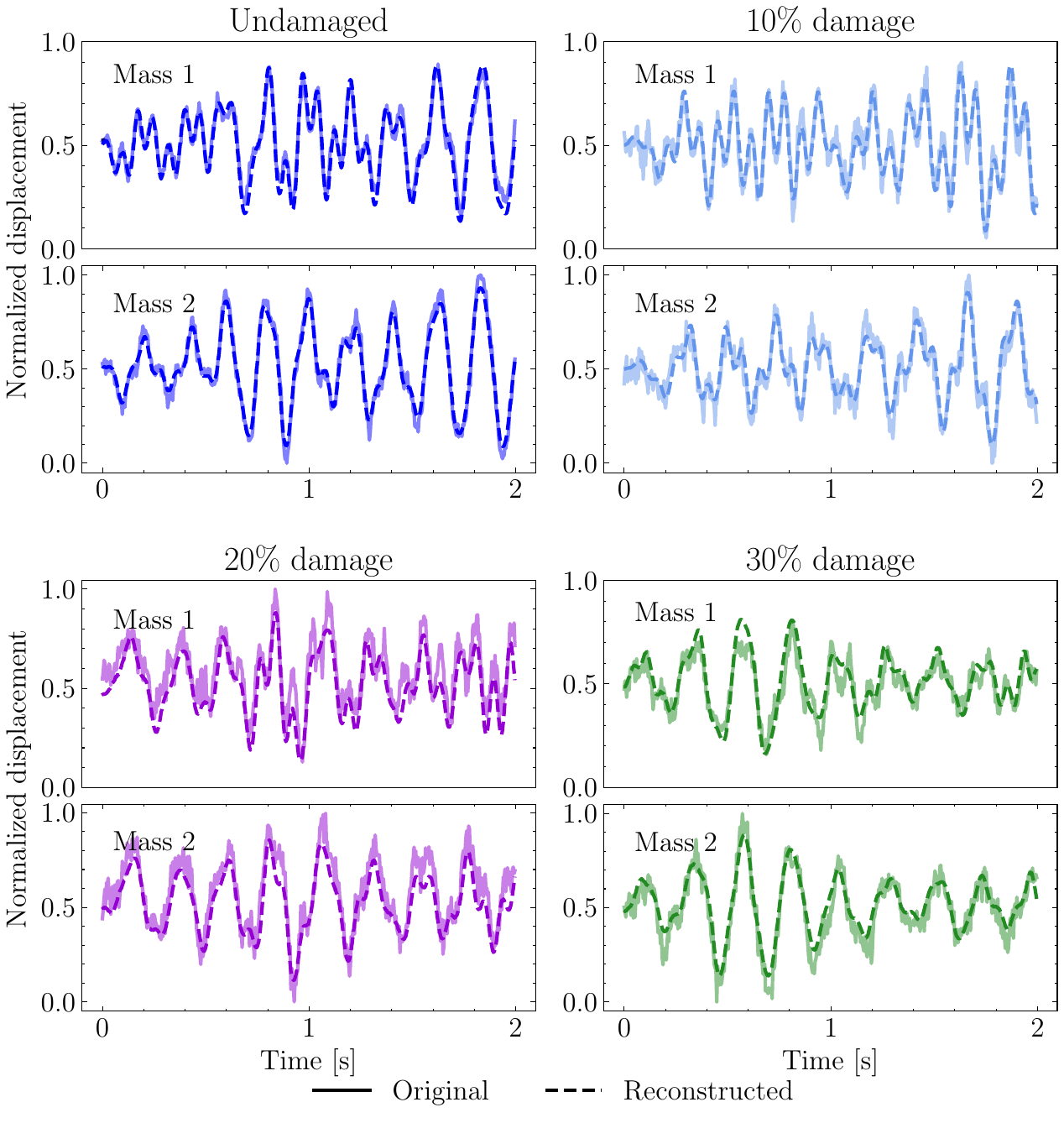}
\caption{Reconstruction of the 2-DOF system response through AE for different damage levels.}
\label{fig:2DOFcubic_ae}
\end{figure}

\begin{figure}[h!]
\centering
\includegraphics[width=1\linewidth]{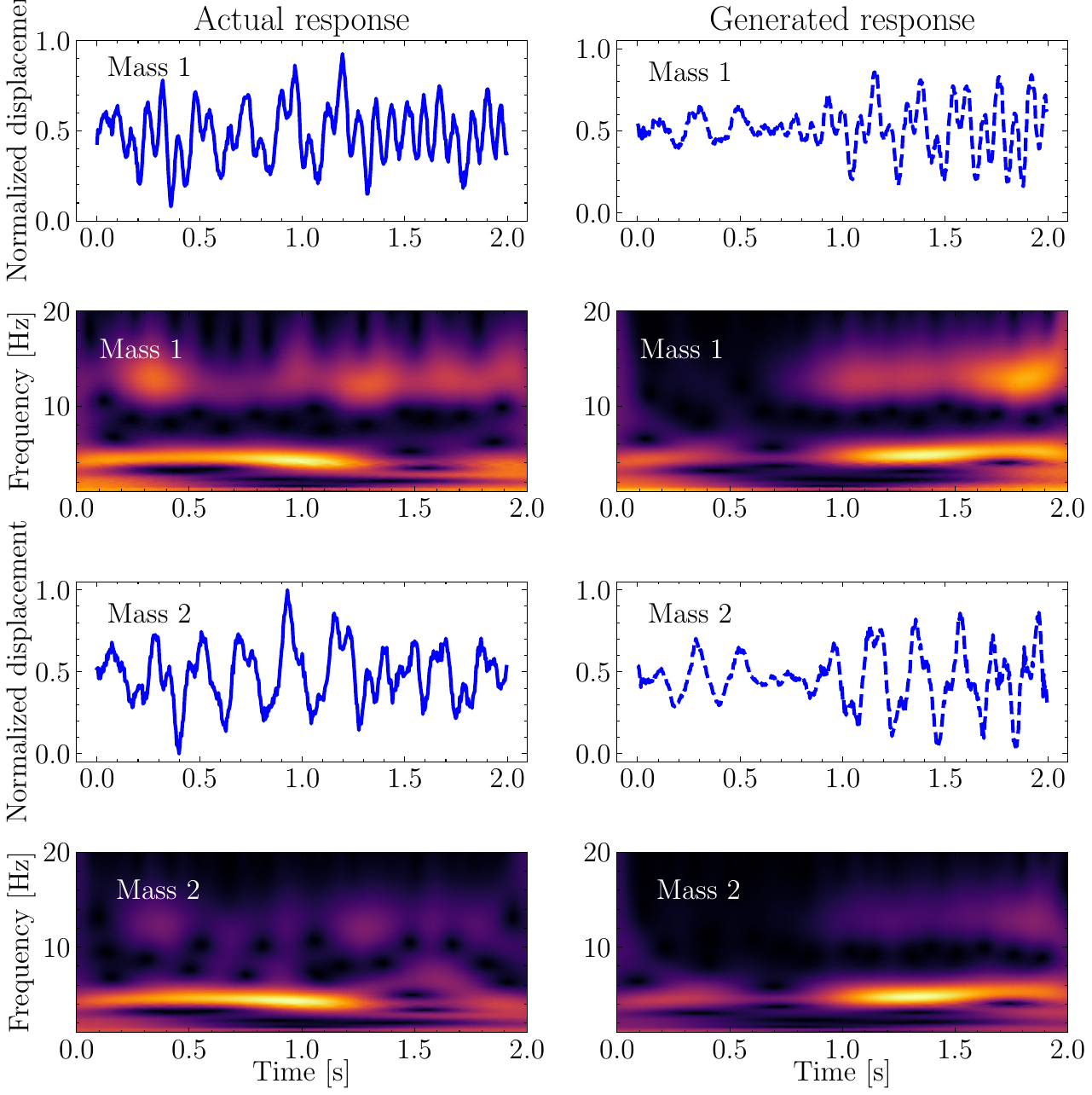}
\caption{Time-history and corresponding time-frequency representation via wavelet transform of the undamaged 2-DOF  system response: actual data and results carried out from the GAN generation block.}
\label{fig:2DOFcubic_gan}
\end{figure}

Although the interaction of the two nonlinear oscillators complicates the overall dynamic behavior, both  architectures confirm their good performances as shown in Figs. \ref{fig:2DOFcubic_ae} and  \ref{fig:2DOFcubic_gan}. As far as the application of the AE is concerned, Fig. \ref{fig:2DOFcubic_ae} shows that the undamaged response of both oscillators is well reconstructed after the training process, whereas the accuracy decreases as expected once the system is damaged. Figure \ref{fig:2DOFcubic_gan} confirms the good performance of GAN since the real response (i.e., the response obtained as the solution of the equations of motion) and the generated response of the undamaged oscillators look similar in the time-frequency domain. This indicates that the discriminator has been successfully trained to recognize the undamaged system response as genuine, while the damaged system response is likely to be classified as fake, with varying degrees of deviation

\begin{figure}[h!]
\centering
\includegraphics[width=1\linewidth]{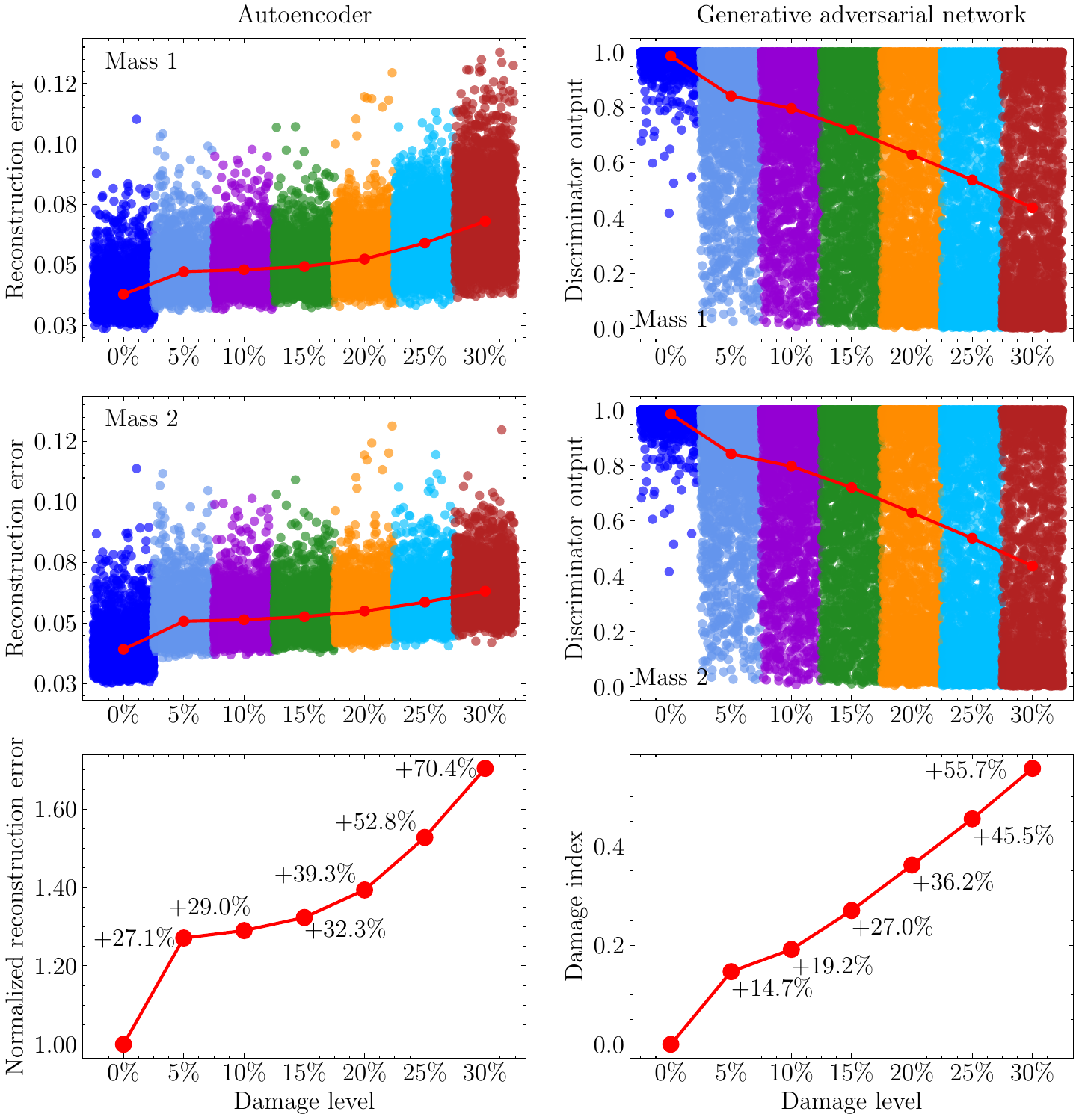}
\caption{Damage sensitivity of the two deep learning architectures for the 2-DOF system: reconstruction loss values obtained by AE and relative variation of its normalized average value, together with discriminator output values obtained by GAN and relative variation of the associated average damage index.}
\label{fig:2DOFcubic_damage}
\end{figure}

The application of both AE and GAN yields the results provided in Fig. \ref{fig:2DOFcubic_damage}. Since both architectures are applied individually to the dynamic response of each mass, Fig. \ref{fig:2DOFcubic_damage} illustrates the reconstruction loss (in terms of MAE) obtained by AE and the discriminator output obtained by GAN for all time series samples of each of the two masses given the damage level. Instead, the trend lines connect the values obtained by averaging the results calculated for both masses in order to enable the detection of damage through the inspection of a single parameter. 

While the considered deep learning architectures displayed a similar performance for the 1-DOF system (see Fig. \ref{fig:1DOFcubic_damage}), Fig. \ref{fig:2DOFcubic_damage} shows that AE is more sensitive to damage than GAN. The lower damage sensitivity of  GAN in this application seems to be associated with  the reduced accuracy of its discriminator  in correctly classifying the undamaged response of the 2-DOF system compared to that of the 1-DOF system. In fact, some discriminator outputs are somewhat far from 1 for the undamaged system, meaning that GAN produces a non-negligible number of false positives (i.e., the discriminator sometimes recognizes as fake the response of the undamaged system, which implies that it detects damage even when none is present). Anyway, Fig. \ref{fig:2DOFcubic_damage} proves that both algorithms enable the early detection of damage and they perform as expected when the damage level increases.

\subsection{Seismic isolator with superelastic hysteresis and negative stiffness}
Health monitoring of seismic isolation systems is crucial to ensure that  the isolated  structures maintain their target performance levels during earthquakes. Consequently, several technical codes and guidelines worldwide require inspections of seismic isolators, both regularly and past seismic events \citep{EN1337,AASHTO2013}. Visual inspections, whenever possible,  can only detect damage or aging effects when they are visible on the surface, and their reliability heavily depends on the inspectors' experience. Therefore, this numerical study investigates the feasibility of using deep learning architectures to detect damage in seismic isolators. An innovative concept of seismic isolator with superelastic hysteresis and negative stiffness proposed by \cite{salvatore2021nonlinear} is here examined. It combines the Bouc-Wen model of hysteresis \citep{bouc1967forced,wen1976method}, a negative stiffness mechanism, and a superelastic model proposed by \cite{charalampakis2018simple}. The equation of motion reads:
\begin{equation}
\begin{aligned}
M\ddot{x} +f_r = F(t),
\label{eq:isolator}
\end{aligned}
\end{equation}
where the total restoring force of the seismic isolation device is given by $f_r = f_i + f_n + f_s $, in which $f_i$ corresponds to the traditional seismic elastomeric isolator restoring force, $f_s$ is the superelastic force and $f_n$ reflects the negative stiffness mechanism. The contribution to the restoring force given by $f_i$ is defined as follows:
\begin{subequations}
\begin{equation}
f_i =C\dot{x}+\alpha K_i x + \left(1-\alpha\right)K_i z,
\end{equation}
\begin{equation}
\dot{z} = \dot{x} [1-(\gamma +\beta \text{sign}(z\dot{x}))]|z| ^n,
\end{equation}
\label{eq:hysteretic}
\end{subequations}
where $z$ is the hysteretic force governed by $\alpha$, which is the ratio between the post-elastic and the stiffness $K_i$ at the origin. Moreover, $\gamma$ and $\beta$ control the shape of the hysteresis and $n$ regulates the smoothness of elastic-to-plastic transition. The contribution to the restoring force given by $f_n$ is defined as follows:
\begin{equation}
f_n = \left(-K_n x + K_3 x^3 \right) \dfrac{ \left[1+\text{sign}\left(x_f-|x|\right)\right] }{2},
\end{equation}
where $K_n$ is the negative linear stiffness, $K_3$ is the positive cubic stiffness, and $x_f$ controls the extent of the negative stiffness force. The last contribution to the restoring force given by $f_s$ is defined as follows \citep{charalampakis2018simple}:
\begin{subequations}
\begin{equation}
\dot{f}_s=  (1-s)K_s\left[\dot{x}-\left|\dot{x}\right|\text{sign}\left(f_s-\beta_s\right)\left( \dfrac{\left|f_s-\beta_s\right|}{Y}\right)^{n_s} \right] + s K_m \dot{x},
\end{equation}
\begin{equation}
\beta_s = K_s\alpha_s\left\{x-\dfrac{f_s}{K_s}+f_t\text{tanh}(a_s x)\left[\dfrac{1+\text{sign}(-x\dot{x})}{2}\right]\right\},
\end{equation}
\begin{equation}
s=\dfrac{\text{tanh}\left[c_s\left(\left|x\right|-x_m\right)\right]+1}{2}, \, f_t = \frac{(2Y - y_sY)}{\alpha_s K_s}, \, a_s = \frac{\tan^{-1}(\tilde{a}_sK_s)} {Y-y_sY}, 
\end{equation}
\label{eq:sma}
\end{subequations}
where $K_s$ and $K_m$ are the initial stiffness during the austenitic phase and the fully martensitic phase, respectively, whereas $Y$ is the yielding force and $\alpha_s$ controls the post-elastic stiffness. The parameter $n_s$ regulates the smoothness of the transition from the initial elastic to the post-elastic phase while $f_t$ and $a_s$ control the twinning hysteresis and super-elasticity and the pinching around the origin along the cycle, respectively.
The complex dynamics resulting from Eqs. \eqref{eq:isolator}-\eqref{eq:sma} is a formidable challenge for testing the ability of AE and GAN to detect damage in nonlinear systems. The system parameters for this application are set to: $M=400$ kg, $C=0.4767$ kg/s, $K_i=1.1$ kN/mm, $K_n=0.5$ kN/mm, $K_3=0.00001$ kN/mm$^3$, $x_u=100$ mm, $x_m=0.7x_u$, $x_f=0.7x_u$, $\alpha=0.2$, $\beta=0.0007$, $c=0.01$, $\gamma=0.00001$, $n=1$, and $n_s=3$. Damage is simulated by reducing both $K_i$ and $K_n$ by the same amount.

Both AE and GAN are assessed through numerical simulations by solving Eq. \eqref{eq:isolator} under random excitation such that $F = -\ddot{x}_b M$, where the peak value of the time-history samples of $\ddot{x}_b$ are scaled to a uniform random value between $0.04g$ and $0.80g$. Table \ref{table:isolator} shows the details about the architecture of both AE and GAN for this numerical application.

\begin{table}[ht]
\centering
\caption{Configuration of the deep learning architectures for damage detection in the seismic isolator.}
\label{table:isolator}
\begin{tabular}{@{}llccccc@{}}
\toprule
\textbf{Type} & \textbf{Component} & \textbf{\makecell{Convolution\\layers}} & \textbf{\makecell{Trainable\\parameters}} & \textbf{\makecell{Kernel\\size}} & \textbf{\makecell{Number\\of filters}} & \textbf{Strides} \\ \midrule
\multirow{2}{*}{AE} & Encoder &  3&38,650  &(20,10,5)  & (50,50,50) &(1,1,1)  \\
            & Decoder &3  & 791,596 &(5,10,20)  &  (50,50,1)& (2,2,1) \\
\midrule
\multirow{2}{*}{GAN}         & Generator & 2 &898  & (2,8) &  (2,1)& (2,2) \\
            & Discriminator &  3& 100,623 &  (8,4,2)&(8,4,2)  & (2,2,1) \\ \bottomrule
\end{tabular}
\end{table}

\begin{figure}[h!]
\centering
\includegraphics[width=1\linewidth]{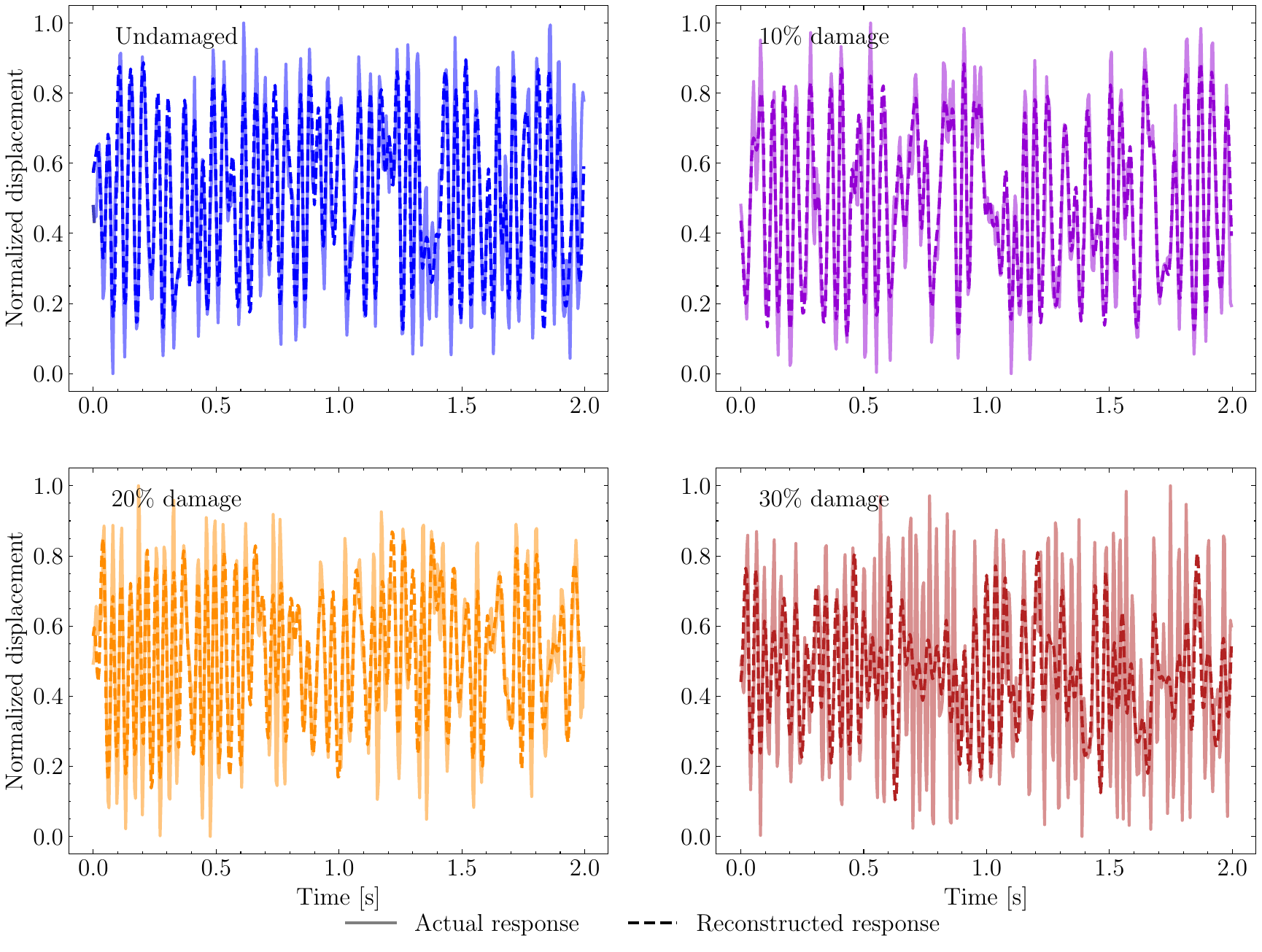}
\caption{Reconstruction of the seismic isolator response through AE for different damage levels.}
\label{fig:isolator_ae}
\end{figure}

\begin{figure}[h!]
\centering
\includegraphics[width=1\linewidth]{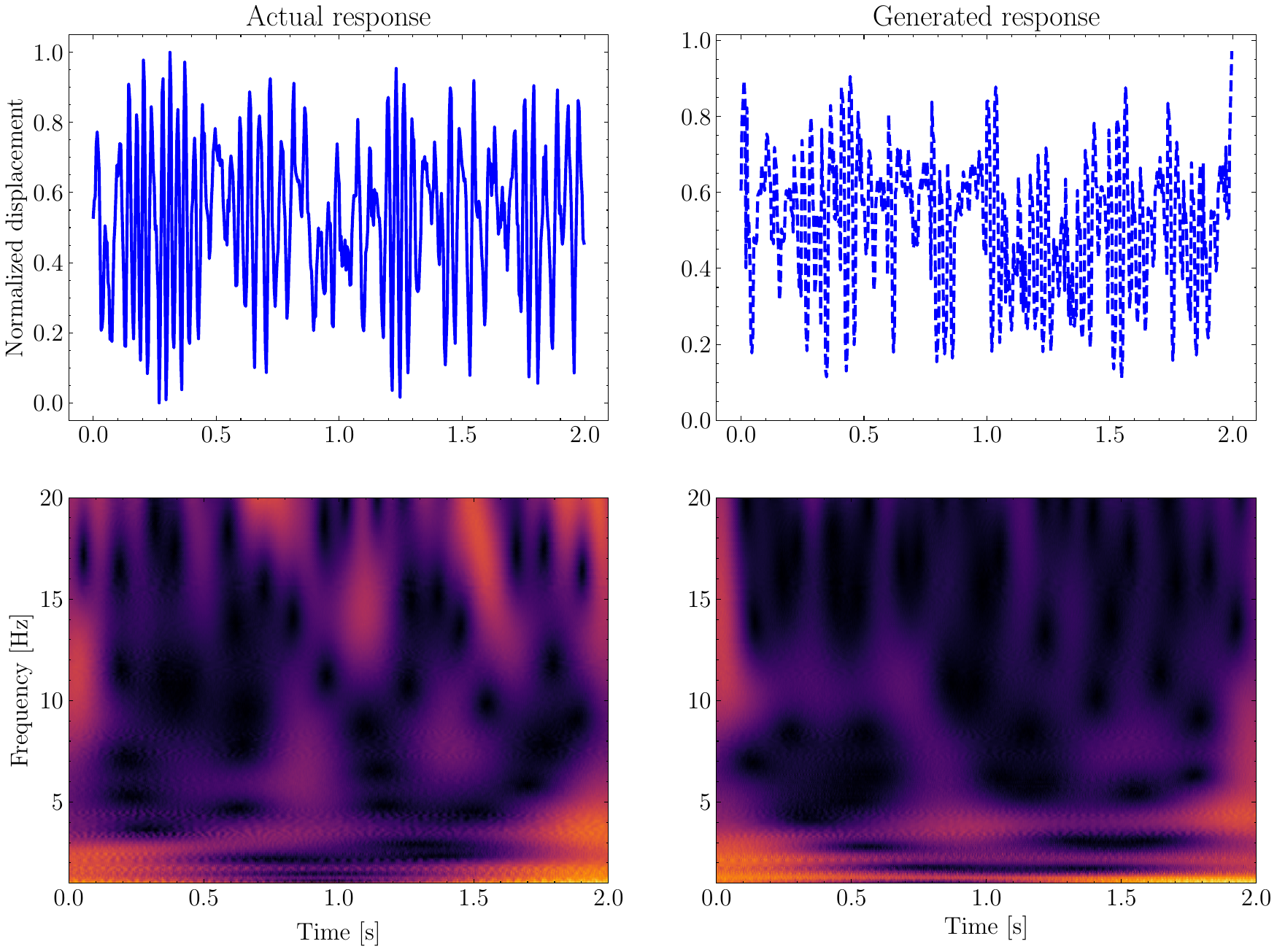}
\caption{Time-history and corresponding time-frequency representation via wavelet transform of the undamaged seismic isolator response: actual data and results carried out from the GAN generation block.}
\label{fig:isolator_gan}
\end{figure}

Representative results carried out from the application of AE and GAN are reported in  Figs. \ref{fig:isolator_ae} and \ref{fig:isolator_gan}, respectively. A close inspection of the results in Fig. \ref{fig:isolator_ae} demonstrates that AE continues to reconstruct the undamaged response of the dynamic system quite satisfactorily while its performance degrades as expected when the level of damage increases. GAN faces some difficulties in this application, as it can be inferred from the comparison in the time-frequency domain between actual and generated response given by Fig. \ref{fig:isolator_gan} (i.e., some frequency components in the actual response are missing in the output produced by the generator). Indeed, the discriminator sometimes mislabels data from the generator as real, even though the comparison with the reference isolator response in the time-frequency domain points out some differences. This is due to the very rich dynamics given by Eqs. \eqref{eq:isolator}-\eqref{eq:sma} and anticipates a lower sensitivity of GAN in detecting damage.

\begin{figure}[h!]
\centering
\includegraphics[width=1\linewidth]{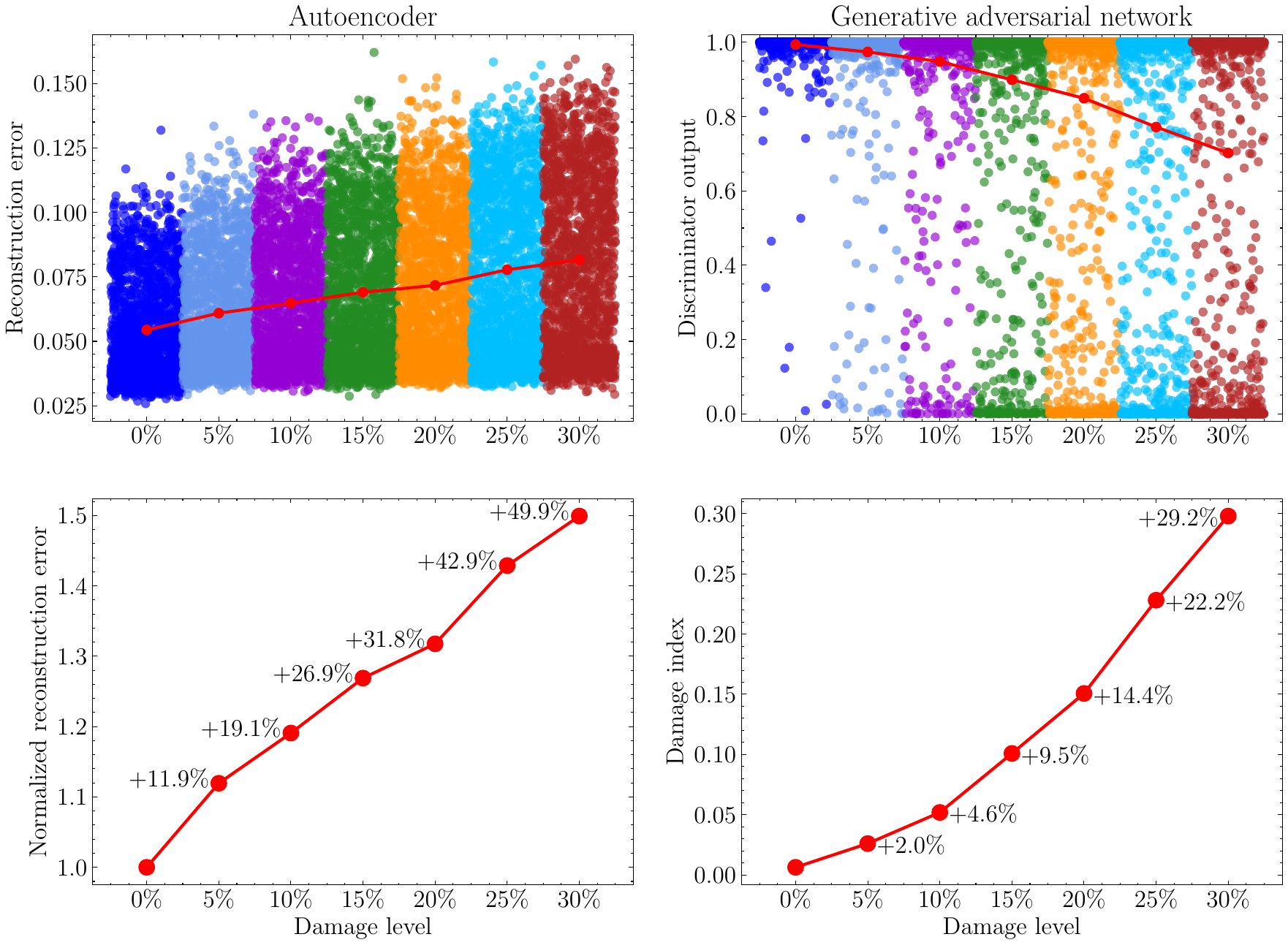}
\caption{Damage sensitivity of deep learning architectures for the seismic isolator: reconstruction loss values obtained by AE and relative variation of its normalized average value, together with discriminator output values obtained by GAN and relative variation of the associated average damage index.}
\label{fig:isolator_damage}
\end{figure}

Results in Fig. \ref{fig:isolator_damage} agree with the evidence drawn from Figs. \ref{fig:isolator_ae} and \ref{fig:isolator_gan}. In fact, Fig. \ref{fig:isolator_damage} shows that the AE properly detects the  presence of damage. The sensitivity of AE in this application remains satisfactory for all damage levels, but it is generally lower compared to Fig. \ref{fig:1DOFcubic_damage} or Fig. \ref{fig:2DOFcubic_damage}. The  presence of  damage in the seismic isolator is also successfully recognized by GAN, provided that the damage severity is large enough (the relative variation of the damage index is around 10\% once the damage level approaches 15\%) while the output is almost inconclusive otherwise. This is mainly because the discriminator tends to produce more false negatives in this application (i.e., too many samples of the damaged response of the isolator are incorrectly classified as true, meaning they are mistakenly considered as representative of  a healthy state).    

%%%%%%%%%%%%%%%%%%%%%%%%%%%%%%%%%%
\section{Experimental application}\label{sec:exp}
\begin{figure}[h!]
\centering
\includegraphics[width=1\linewidth]{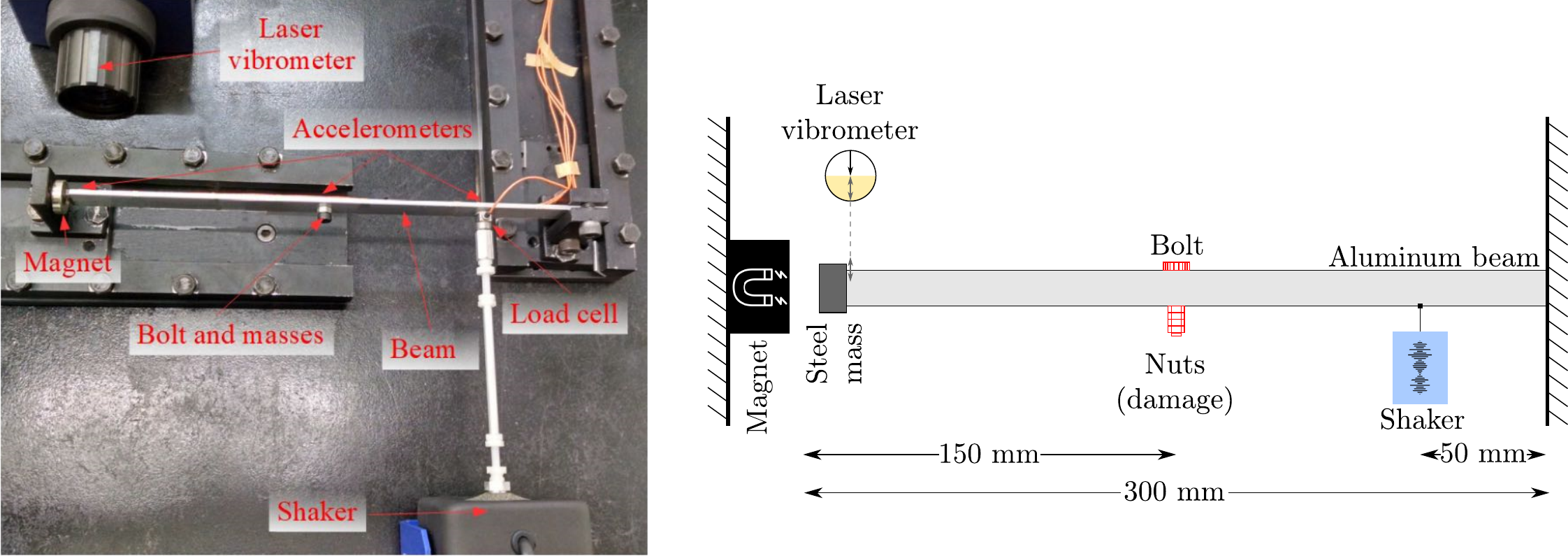}
\caption{Layout of the experimental test consisting of a cantiliver beam and a magnet close to the tip mass \citep[][reprinted with permission]{villani2019damageExp}.}
\label{fig:magnolia_setup}
\end{figure}

Both AE and GAN are finally tested using real data from a laboratory experiment reported by \cite{shiki2017application} and \cite{villani2019damageExp}. The experimental setup is shown in Fig. \ref{fig:magnolia_setup} and is composed by a cantilever aluminum beam having dimensions of 300 mm $\times$ 19 mm $\times$ 3.2 mm with a small steel mass attached to its free end that interacts with a neodymium magnet positioned at a distance of 2 mm. A bolted connection provided with up to four nuts is positioned 150 mm from the free end to simulate the presence of damage. Considering the mass of each nut and varying the number of nuts, the damage levels turn out to be equal to 2\%, 4\% and 6\%. An electrodynamic shaker is placed 50 mm away from the clamped end of the beam whereas a laser vibrometer measures the velocity of the free end. A hardening-type nonlinearity originates in this magneto-elastic dynamic system from the interaction between the permanent magnet and the steel mass at the beam free end. While chirp and stepped sine were used as excitation by \cite{shiki2017application} and \cite{villani2019damageExp}, the present experimental application deals with output-only damage detection under random vibrations. Accordingly, the measured system response under band-pass filtered white noise is considered. The filter bandwidth in the experiment was set to 10-420 Hz whereas the input excitation levels were equal to 0.01 VRMS, 0.05 VRMS, 0.10 VRMS, 0.11 VRMS, 0.12 VRMS, 0.13 VRMS, 0.14 VRMS and 0.15 VRMS. 

Once again, the system dynamic response under different (unknown) excitation levels is taken into account to assess the effectiveness of the deep learning architectures in detecting damage, regardless of the experienced level of nonlinearity. The configuration parameters  of AE and GAN adopted in such experimental application are given in Tab. \ref{table:magnolia}.

\begin{table}[ht]
\centering
\caption{Configuration of the deep learning architectures for damage detection in the magneto-elastic system.}
\label{table:magnolia}
\begin{tabular}{@{}llccccc@{}}
\toprule
\textbf{Type} & \textbf{Component} & \textbf{\makecell{Convolution\\layers}} & \textbf{\makecell{Trainable\\parameters}} & \textbf{\makecell{Kernel\\size}} & \textbf{\makecell{Number\\of filters}} & \textbf{Strides} \\ \midrule
\multirow{2}{*}{AE} & Encoder &  4&1,283,525  &(20,100,50,25)  & (100,100,50,25) &(1,1,1,1)  \\
            & Decoder &3  & 527,679            &(50,50,20)         &  (50,100,1).       & (1,2,2) \\
\midrule
\multirow{2}{*}{GAN}         & Generator & 3 &414,629  & (10,20,2) &  (100,200,1)& (1,2,2) \\
            & Discriminator &  4& 2,357,201 &  (3,50,15,10)&(200,200,100,50)  & (1,1,1,1) \\ \bottomrule
\end{tabular}
\end{table}

\begin{figure}[h!]
\centering
\includegraphics[width=1\linewidth]{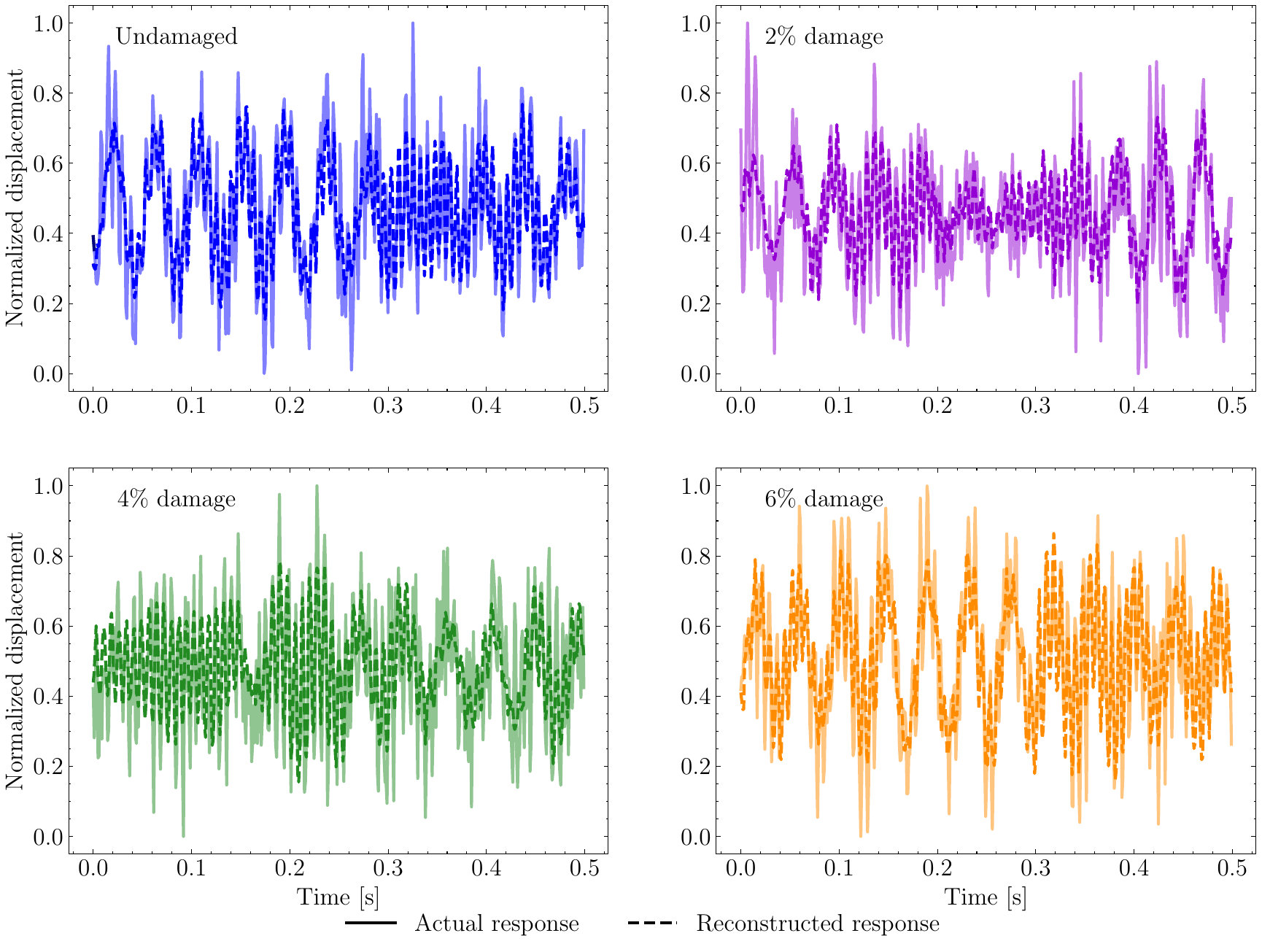}
\caption{Reconstruction of the magneto-elastic system response through AE for different damage levels.}
\label{fig:magnolia_ae}
\end{figure}

\begin{figure}[h!]
\centering
\includegraphics[width=1\linewidth]{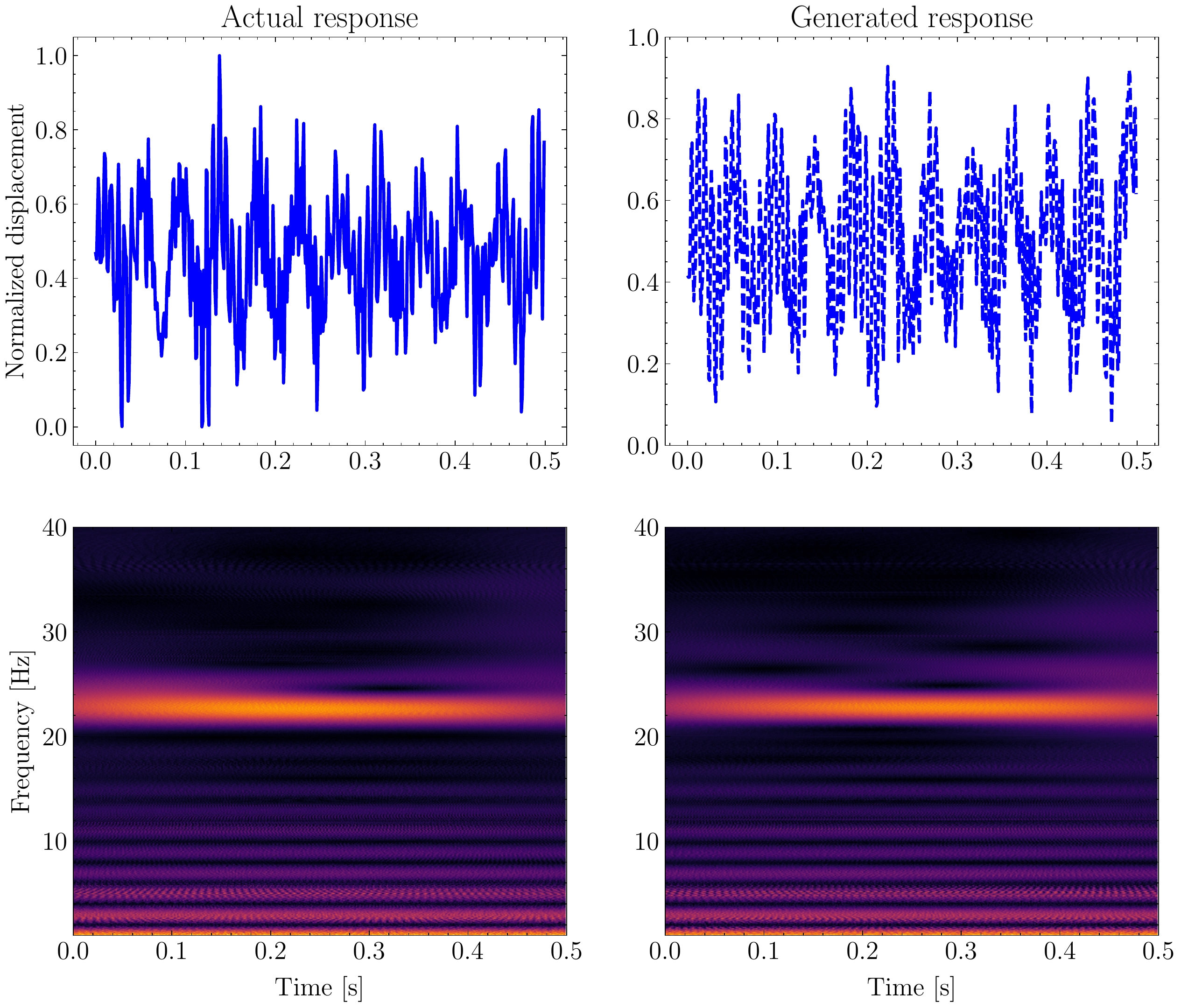}
\caption{Time-history response and corresponding time-frequency representation via wavelet transform of the undamaged magneto-elastic system: actual data and results carried out from the GAN generation block.}
\label{fig:magnolia_gan}
\end{figure}

\begin{figure}[h!]
\centering
\includegraphics[width=1\linewidth]{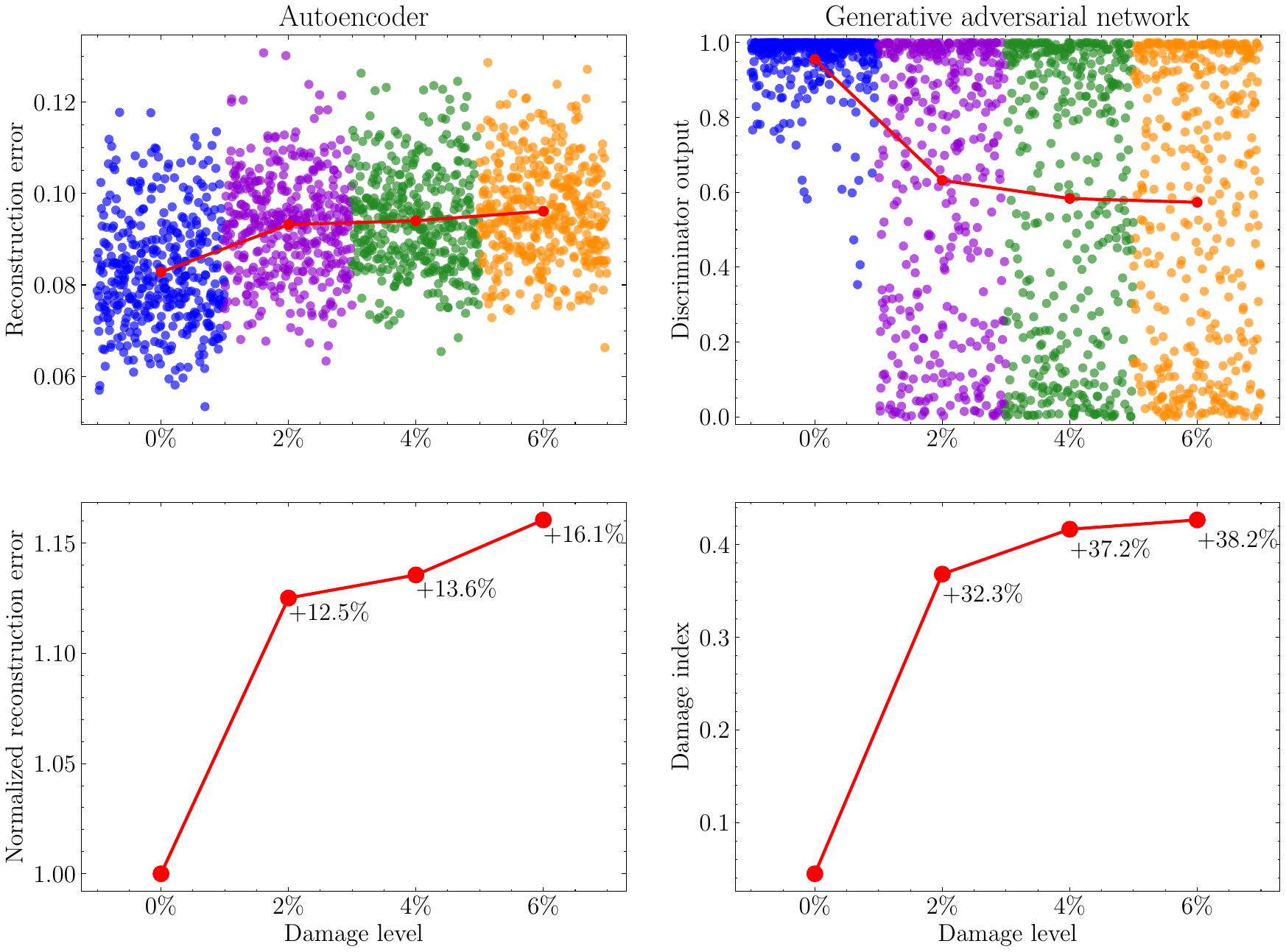}
\caption{Damage sensitivity of deep learning architectures for the magneto-elastic system: reconstruction loss values obtained by AE and relative variation of its normalized average value, together with discriminator output values obtained by GAN and relative variation of the associated average damage index.}
\label{fig:magnolia_damage}
\end{figure}

Figures \ref{fig:magnolia_ae} and \ref{fig:magnolia_gan} show the performances of both architectures after training while Fig. \ref{fig:magnolia_damage} demonstrates their effectiveness in detecting damage. On the one hand, the reconstruction error of AE displays an abrupt increment as soon as a small damage occurs. On the other hand, a growing number of time series data is classified as fake by the discriminator of GAN once a small damage occurs because they do no longer represent the dynamics of a healthy system. The very good sensitivity of both architectures for the lowest damage level of 2\% confirms their ability in early damage detection. In particular, the damage index based on the discriminator loss  highlights the exceptional performance of GAN in recognizing the onset of damage in this experimental application. However, the relative variation of the normalized reconstruction loss and that of the damage index based on the discriminator loss for damage levels of 4\% and 6\% are fairly close to the corresponding values observed for the minimum damage level equal to 2\%. Ultimately, this experimental application further confirms that both AE and GAN are able to detect damage without a priori information about the system or the applied dynamic loads, and irrespective of the degree of nonlinearity. 

%%%%%%%%%%%%%%%%%%%%%%%%%%%%%%%%%%
\section{Conclusions} \label{sec:end}
The original contribution of this work lies in the first comprehensive study on using deep learning architectures for unsupervised, data-driven, vibration-based damage detection in nonlinear dynamic systems. The performance of both AE and GAN was deeply examined, considering different nonlinear behaviors. The excitation scenarios involve random excitation with varying intensity. Numerical investigations and experimental applications have been conducted under the assumption that neither the system nor the excitation are known. The following conclusions can be drawn from the present extensive study on different systems and datasets.
\begin{itemize}
\item Both AE and GAN are able to detect the onset of damage in different nonlinear dynamic systems. They also show consistent results upon increasing damage levels.
\item On the one hand, it seems that AE displays a more consistent performance than GAN. GAN did not consistently detect the lowest damage level; however, it can exhibit exceptional damage sensitivity.
\item The AE architecture definition proved easier than configuring GAN. Training AE was also less time consuming than GAN. Conversely, the main advantage of the GAN discriminator lies in its output being interpretable as damage index since it is bounded between two limit values corresponding to healthy and damaged state. While GAN is applied to detect the occurrence of damage, it can be possibly used for other tasks, such as synthetic data generation, uncertainty quantification, and reliability assessment. 
\end{itemize}
In addition to extending both numerical and experimental applications, some issues should be addressed in future efforts, as listed below.
\begin{itemize}
\item It is known that the effects attributable to environmental conditions can obscure  those due to damage. Therefore, understanding whether deep learning architectures can detect effectively damage under changing environmental conditions without preliminary data processing  is still an open issue.
\item Both AE and GAN were employed to establish whether the nonlinear dynamic system was damaged or not (i.e., novelty detection). Damage localization and quantification were not addressed. Further studies should aim to understand whether the output of such deep learning architectures can be eventually correlated with the damage position and/or magnitude. 
\end{itemize}

\section*{Acknowledgments} \label{sec:ack}
The authors would like to thank Americo Cunha (Rio de Janeiro State University, Brazil) and Samuel Da Silva (S\~{a}o Paulo State University, Brazil) for sharing the experimental data. 

This work was partially carried out within the RETURN Extended Partnership funded by the European Union Next-GenerationEU (National Recovery and Resilience Plan -- NRRP, Mission 4, Component 2, Investment 1.3 -- D.D. 1243 2/8/2022, PE0000005).

%%%%%%%%%%%%%%%%%%%%%%%%%%%%%%
\bibliographystyle{spbasic}
\bibliography{refs}

\end{document}